%% file: main.tex
\documentclass[letterpaper]{article} 
\usepackage[draft]{aaai2026}  
\usepackage{times}  
\usepackage{helvet}  
\usepackage{courier}  
\usepackage[hyphens]{url}  
\usepackage{graphicx} 
\usepackage{natbib}  
\usepackage{caption} 
\input{tex-src/packages}
\input{tex-src/macros}
\input{tex-src/math-commands}

\title{Guess or Recall? Training CNNs to Classify and Localize Memorization in LLMs}
\author {
    Jérémie Dentan\textsuperscript{\rm 1},
    Davide Buscaldi\textsuperscript{\rm 1, 2},
    Sonia Vanier\textsuperscript{\rm 1}
}
\affiliations {
    \textsuperscript{\rm 1}LIX (École Polytechnique, IP Paris, CNRS)\\
    \textsuperscript{\rm 2}LIPN (Université Sorbonne Paris Nord)\\
    jeremie.dentan@polytechnique.edu, davide.buscaldi@polytechnique.edu, sonia.vanier@polytechnique.edu
}

\begin{document}

\pagestyle{plain}
\thispagestyle{plain} 
\pagenumbering{arabic}

\maketitle
\input{sections/00_abstract}
\input{sections/01_introduction}
\input{sections/02_related_works}
\input{sections/03_methodology}
\input{sections/04_empirical_results}
\input{sections/05_localizing_memorization}
\input{sections/06_concluding_remarks}

\bibliography{bib/custom}

\input{sections/a_01_implementation_details}
\input{sections/a_02_details_parametrization}
\input{sections/a_03_additional_figures}

\end{document}

%% file: tex-src/packages.tex
\usepackage{algorithm}
\usepackage{algorithmic}

\usepackage{newfloat}
\usepackage{listings}
\DeclareCaptionStyle{ruled}{labelfont=normalfont,labelsep=colon,strut=off} 
\floatstyle{ruled}
\newfloat{listing}{tb}{lst}{}
\floatname{listing}{Listing}

\usepackage{times}
\usepackage{latexsym}

\usepackage{subcaption}
\usepackage{adjustbox}
\usepackage{amsfonts}
\usepackage{amsmath}
\usepackage{booktabs}
\usepackage{stmaryrd}
\usepackage{mathtools}
\usepackage{amssymb}
\usepackage{cprotect}

%% file: tex-src/macros.tex
\newcommand{\hdelta}{\bm \delta}
\newcommand{\hlambda}{\bm \lambda}
\newcommand{\hgamma}{\bm \gamma}
\DeclareMathOperator{\average}{average}
\DeclareMathOperator{\relu}{ReLU}

\makeatletter
\if T\showauthors@on
    \newcommand{\safelink}[2]{\url{#1}}
    \newcommand{\acknowledge}[1]{\section*{Acknowledgements} #1}
\else
    \newcommand{\safelink}[2]{[Link placeholder for review: #2]}
    \newcommand{\acknowledge}[1]{}
\fi
\makeatother

\makeatletter
\def\blfootnote{\gdef\@thefnmark{}\@footnotetext}
\makeatother

%% file: tex-src/math-commands.tex
\usepackage{amsmath,amsfonts,bm}

\def\eqref#1{equation~\ref{#1}}

\def\1{\bm{1}}

\def\mA{{\bm{A}}}
\def\mB{{\bm{B}}}
\def\mC{{\bm{C}}}
\def\mD{{\bm{D}}}

\DeclareMathAlphabet{\mathsfit}{\encodingdefault}{\sfdefault}{m}{sl}
\SetMathAlphabet{\mathsfit}{bold}{\encodingdefault}{\sfdefault}{bx}{n}



%% file: sections/00_abstract.tex
\begin{abstract}

Verbatim memorization in Large Language Models (LLMs) is a multifaceted phenomenon involving distinct underlying mechanisms. We introduce a novel method to analyze the different forms of memorization described by the existing taxonomy. Specifically, we train Convolutional Neural Networks (CNNs) on the attention weights of the LLM and evaluate the alignment between this taxonomy and the attention weights involved in decoding. We find that the existing taxonomy performs poorly and fails to reflect distinct mechanisms within the attention blocks. We propose a new taxonomy that maximizes alignment with the attention weights, consisting of three categories: memorized samples that are \textit{guessed} using language modeling abilities, memorized samples that are \textit{recalled} due to high duplication in the training set, and \textit{non-memorized} samples. Our results reveal that few-shot verbatim memorization does not correspond to a distinct attention mechanism. We also show that a significant proportion of extractable samples are in fact guessed by the model and should therefore be studied separately. Finally, we develop a custom visual interpretability technique to localize the regions of the attention weights involved in each form of memorization.

\end{abstract}

\begin{links}
    \link{Code}{https://github.com/orailix/cnn-4-llm-memo}
\end{links}

%% file: sections/01_introduction.tex
\section{Introduction}

\input{figures/teaser/teaser}

\blfootnote{Preprint. This paper has been accepted for publication at AAAI-26. See the published version for the copyright notice.}

Large Language Models (LLMs) are known to memorize a significant portion of their training data, raising legal and ethical challenges \citep{zhang_understanding_2017, mireshghallah_empirical_2022, carlini_quantifying_2023}. Recently, \citet{prashanth_recite_2024} view memorization as a multifaceted phenomenon and propose a taxonomy of memorized samples. They focus on training set samples that are \textit{32-extractable}: when prompted with the first 32 tokens of a sequence (the \textit{prefix}), the model outputs exactly the next 32 tokens (the \textit{suffix}). Samples are categorized into four classes: \textit{Non-memorized, Recite, Reconstruct, and Recollect} (defined in Figure~\ref{fig:teaser}). This taxonomy aims to capture the different mechanisms underlying each type of memorization and is motivated by high-level features such as perplexity and token frequency. The authors show that each category exhibits distinct and consistent high-level characteristics across memorized samples. 

In this paper, we take a step further by investigating whether these different forms of memorization can be observed at the level of the model's attention weights. Since the samples are 32-extractable, there must be causal links between the prefix and the suffix that enable the exact decoding of the latter. We therefore analyze the attention weights to uncover such links. Building on the taxonomy of \citet{prashanth_recite_2024}, we ask whether the nature of these causal links differs across the categories. To identify distinctive patterns in attention weights across heads and layers, we trained CNNs to classify them according to the taxonomy of \citet{prashanth_recite_2024} (see Figure~\ref{fig:teaser}; further details are discussed in later sections). The CNNs revealed that the classes proposed by \citet{prashanth_recite_2024} do not align well with the attention weights, as evidenced by frequent misclassifications. For example, Figure~\ref{fig:example_attention_weights} shows a sample that should clearly belongs to the \textit{Reconstruct} class but is labeled \textit{Recollect}. Similarly, we observed many \textit{Recite} samples whose attention weights closely resemble those of \textit{Reconstruct}, as reflected in the CNNs’ frequent misclassifications.

These misclassifications are problematic, as they suggest that this taxonomy does not reflect the underlying causal mechanisms between the prefix and the suffix in memorized samples. For example, we observed that \textit{Recollect} does not represent a truly distinct form of memorization; rather, these samples should be reassigned to either \textit{Recite} or \textit{Reconstruct}, depending on which group their attention weights most closely resemble (see Figure \ref{fig:sankey}). This distinction is crucial, as it prompts a reevaluation of what models are capable of memorizing: few-shot memorization of samples observed only few times may be illusory. This aligns with the findings of \citet{huang_demystifying_2024} and calls for a reconsideration of current approaches to mitigate memorization.

To address the limitations of the taxonomy proposed by \citet{prashanth_recite_2024}, we introduce a new, simpler, data-driven taxonomy of memorized samples. We developed a protocol to systematically explore multiple candidate taxonomies and evaluate their alignment with attention weights by measuring the performance of CNNs trained to classify attention weights under each taxonomy. Based on this protocol, we propose a new taxonomy that ranks highest in our benchmark and accurately captures distinct mechanisms in the attention weights. It comprises three classes defined in Figure \ref{fig:teaser}. \textit{Non-Memorized} class is similar to that of \citet{prashanth_recite_2024}. \textit{Guess} captures samples where most suffix tokens can be inferred from the prefix. It includes ROUGE-based rules to improve \textit{Reconstruct} class in \citet{prashanth_recite_2024}. All remaining samples are assigned to \textit{Recall}, as we did not observe distinct subgroups that would justify further splitting. See examples in Figure~\ref{fig:example_attention_weights}.

Using a well-designed taxonomy such as ours is crucial for accurately studying memorization. For instance, \citet{stoehr_localizing_2024} identified an attention head in the lower layers that is highly correlated with memorization. Our results suggest that these lower layers primarily contribute to memorizing \textit{Guess} samples. While their finding is valid, it is unlikely to generalize to all forms of memorization. To demonstrate the benefits of studying \textit{Guess} and \textit{Recall} as distinct forms of memorization, we developed a custom technique to identify the regions of the attention weights that play a significant role for each case. Our results show that \textit{Guess} samples exhibit high activations in the lower layers of the model, consistent with the observations of \citet{stoehr_localizing_2024}. We also show that \textit{Recall} relies on short-range interactions between neighboring tokens in the upper layers, corroborating the findings of \citet{huang_demystifying_2024} and \citet{menta_analyzing_2025}. These fine-grained localization results are made possible by our taxonomy, which disentangles truly distinct memorization mechanisms.

\input{figures/sankey_diag/sankey}

\paragraph{Our contributions can be summarized as follows}

\begin{itemize}
    \item We propose a new method to analyze the role of attention blocks in memorization, by training CNNs on attention weights.
    \item We benchmark the alignment of various taxonomies with attention weights, including \citet{prashanth_recite_2024}'s.
    \item We introduce a new data-driven taxonomy that addresses the limitations of previous ones and maximizes alignment with attention weights: \textit{Guess, Recall, Non-memorized}.
    \item We develop a method to interpret the CNNs and localize the regions of the LLM involved in memorization. Our conclusions bridge the gap between the results of \citet{stoehr_localizing_2024}, \citet{huang_demystifying_2024}, and \citet{menta_analyzing_2025}.
\end{itemize}

\input{figures/example_attention_weights/attention_weights}

%% file: figures/teaser/teaser.tex
\begin{figure}[t!]
    \centering
    \includegraphics[width=\columnwidth]{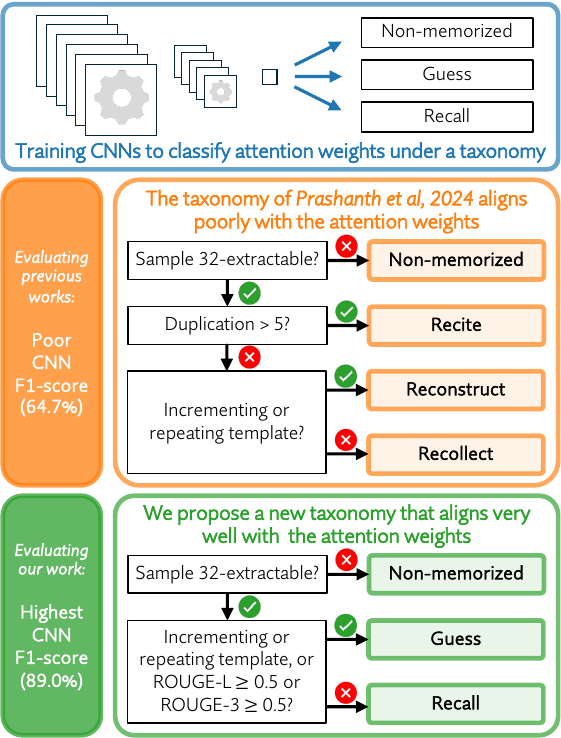}
    \caption{To evaluate a taxonomy of memorized samples, we train CNNs to classify attention weights under this taxonomy. The existing taxonomy yields poor performance. Our new, simpler taxonomy aligns much closely with the attention mechanism involved in data regurgitation.}
    \label{fig:teaser}
\end{figure}

%% file: figures/sankey_diag/sankey.tex
\begin{figure}[t!]
    \centering
    \includegraphics[width=\columnwidth]{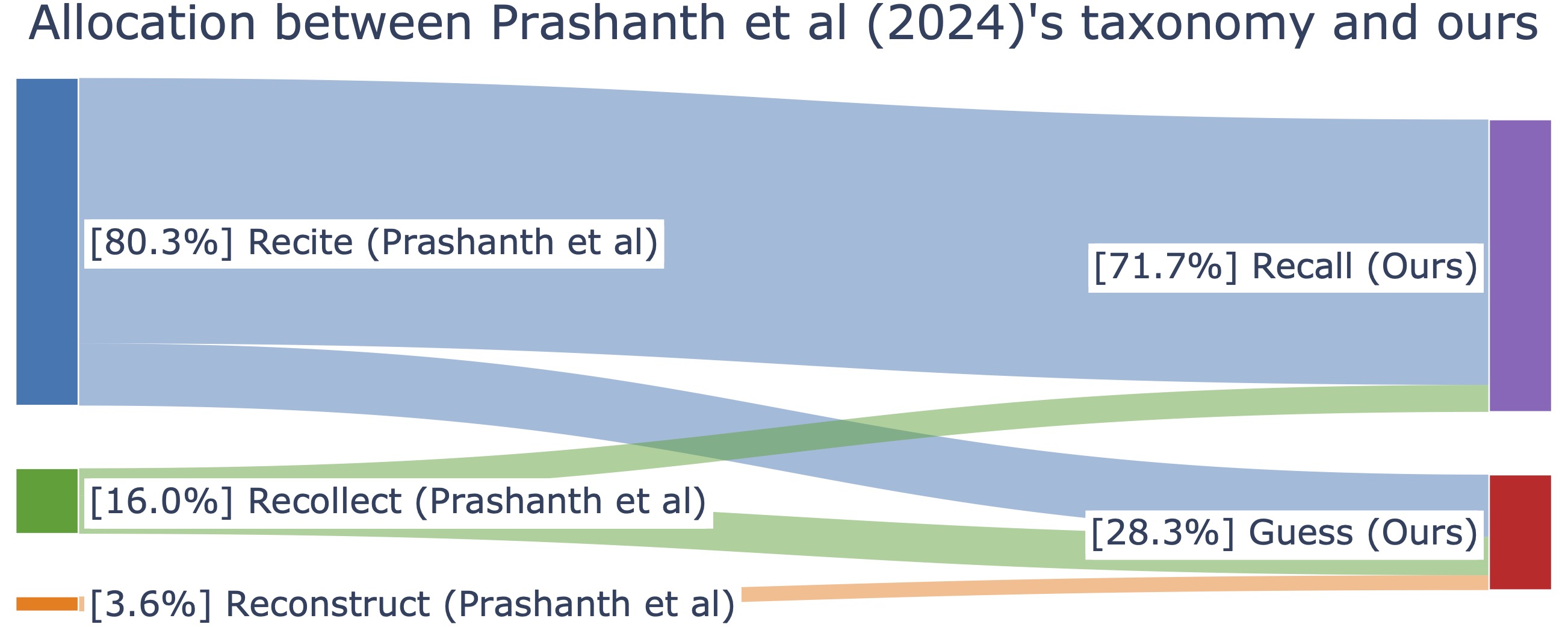}
    \caption{Comparison of samples' labels between \citet{prashanth_recite_2024}'s taxonomy and ours. \textit{Guess} class is broader than \textit{Reconstruct}, including all samples where the suffix largely predictable from the prefix, which exhibit similar attention weights. We omit non-memorized samples here.}
    \label{fig:sankey}
\end{figure}

%% file: figures/example_attention_weights/attention_weights.tex
\begin{figure*}[t!]
    \centering
    \includegraphics[width=\columnwidth]{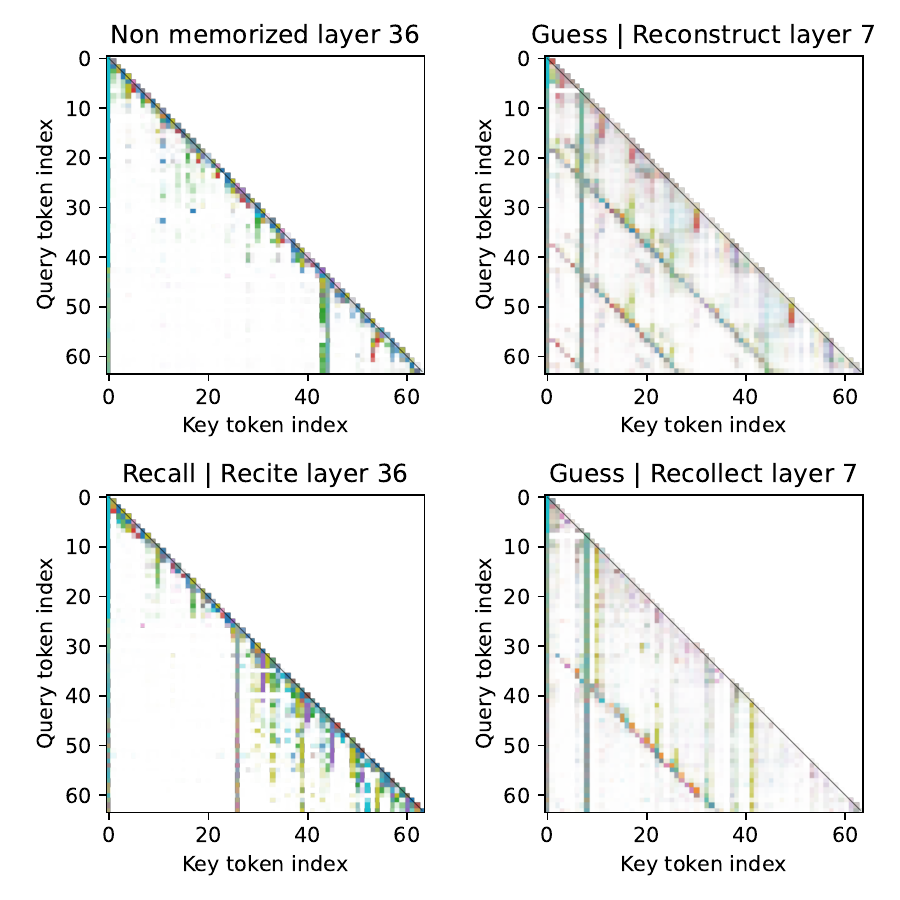}
    \hspace{12pt}
    \includegraphics[width=\columnwidth]{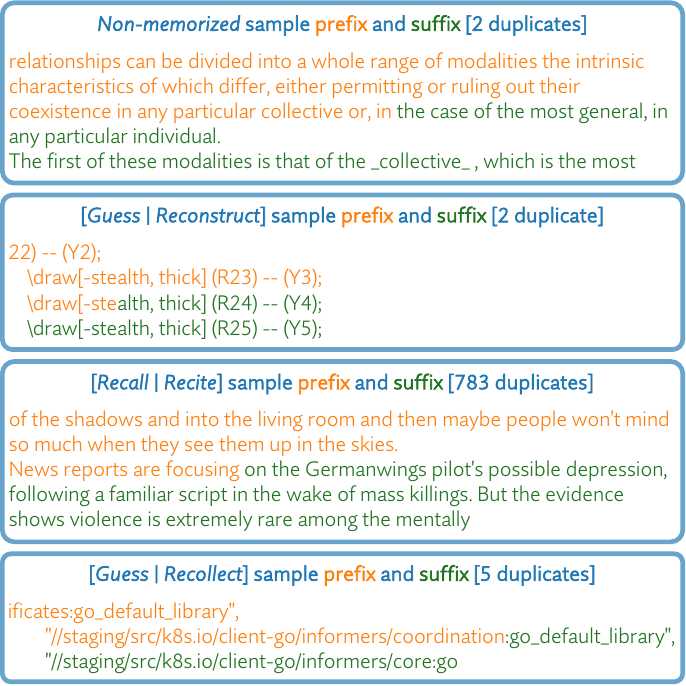}
    \caption{Sample attention weights and their corresponding 64-token text snippets. Labels like [\textit{Guess} $|$ \textit{Reconstruct}] indicate the sample’s class in our taxonomy (left) and in that of \citet{prashanth_recite_2024} (right). The intensity of each color in the matrices represents the attention of a different head. The second and fourth samples exhibit similar patterns in lower-layer attention and are both classified as \textit{Guess} in our taxonomy, though assigned to different classes by \citet{prashanth_recite_2024}.}
    \label{fig:example_attention_weights}
\end{figure*}

%% file: sections/02_related_works.tex
\section{Background and related works}

\subsection{Verbatim memorization in LLMs}

Training data memorization is a broad phenomenon that affects many types of models \citep{fredrikson_privacy_2014, mahloujifar_membership_2021, carlini_extracting_2021, carlini_extracting_2023, dentan_reconstructing_2024}. In this work, we focus on \textit{verbatim memorization} in LLMs: a sample is considered memorized if it is \textit{extractable} from a prompt using greedy decoding \citep{carlini_extracting_2021, carlini_quantifying_2023, yu_bag_2023, nasr_scalable_2025, zhang_extending_2025, chen_multi-perspective_2024}. This setting differs from membership inference \citep{shokri_membership_2017, carlini_membership_2022} and counterfactual memorization \citep{feldman_what_2020, zhang_counterfactual_2023}, which are more common with non-generative models. Although \textit{extractability} has some limitations \citep{ippolito_preventing_2023}, it is fast to compute and widely used across different scenarios \citep{biderman_emergent_2023, lee_language_2023}.

\subsection{Which Samples Are Memorized by LLMs?}

Memorized samples mostly consist of samples that are difficult for the model to represent during training \citep{dentan_predicting_2025, feldman_what_2020, zhang_counterfactual_2023}. The extreme case consists of random sequences of tokens, which cannot be represented using language modeling abilities and are therefore very likely to be memorized \citep{meeus_copyright_2024}. However, most memorized samples are non-random natural language sentences, and \citet{carlini_quantifying_2023} have shown that LLMs memorize up to 1\% of their training data. \citet{prashanth_recite_2024} proposed a taxonomy of memorized samples, presented in Figure \ref{fig:teaser}. Despite its limitations discussed in this paper, their framework captures the diversity of memorization forms and their underlying mechanisms.

\subsection{Localizing memorization in LLMs}

Verbatim memorization is deeply entangled with the general language abilities of LLMs \citep{huang_demystifying_2024}. This entanglement explains why memorization and generalization can reinforce each other \citep{feldman_does_2020}, and why techniques for localizing memorization resemble those used to localize factual knowledge in models \citep{meng_locating_2022}. \citet{stoehr_localizing_2024} observed that memorized samples exhibit larger gradients in the lower layers and identified a specific attention head in these layers strongly associated with memorization. \citet{huang_demystifying_2024} showed that certain token sequences in the prefix act as \textit{triggers}. Their representations in the lower layers encode influential tokens in the suffix, and the model fills in the gaps using language modeling abilities. Unlike knowledge of a single fact, verbatim memorization of a full paragraph cannot be reduced to a single encoding at a specific point in the model. Instead, it is distributed across numerous triggers entangled with the model’s general-purpose capabilities. Finally, \citet{menta_analyzing_2025} demonstrated that deactivating attention blocks in the highest layers can reduce verbatim memorization while preserving performance.

Thus, there appears to be a gap between the role of the early layers highlighted by \citet{huang_demystifying_2024} and \citet{stoehr_localizing_2024}, and that of the final layers successfully used by \citet{menta_analyzing_2025}. Our experiments complement these works and resolve this gap by analyzing the role of attention blocks separately for each form of memorization.

%% file: sections/03_methodology.tex
\section{Taxonomy Benchmark: Methodology} \label{sec:benchmark_methodology}

\subsection{Training CNNs on attention weights} \label{sec:bechnmark_methodology_cnns}

We consider \textit{samples} of 64 contiguous tokens from The Pile \citep{gao_pile_2020}, which is the training set used for Pythia models \citep{biderman_pythia_2023}. The choice of Pythia and The Pile was determined by the need to know the complete set of training data, information that is omitted for most available models. We consider the complete set of 32-extractable samples from Pythia, provided by \citet{biderman_emergent_2023} and refined by \citet{prashanth_recite_2024}, which enables us to make a fair comparison with their taxonomy.

For each sample $s$, we examine the \textit{attention weight} at layers $l$ and attention head $h$, denoted by $\mA^h_l[s] \in \mathbb{R}^{64 \times 64}$. It is the triangular matrix containing at position $(i,j)$ the attention between \textit{query} token $i$ and \textit{key} token $j$ for $0 \le j \le i \le 63$. See examples in Figure \ref{fig:example_attention_weights}. Some aspects are easy to interpret. The vertical bars spanning from the main diagonal to the bottom axis correspond to line breaks, which are crucial for locating a token’s position in the text and therefore receive high attention. Similarly, diagonal lines in the second and fourth sample correspond to approximate repetition of subsequences, which exhibit strong mutual attention.

The most visible patterns in the attention weights are effectively translation-invariant: translating a subsequence of tokens also translates the underlying patterns. For example, we observed that moving an idiomatic expression modifies individual attention weights, but preserves the strong attention between its tokens. Similarly, adding tokens to a repeated sequence shifts the diagonal line in the attention weights without changing its nature. CNNs therefore appear to be a good choice of architecture, well suited for translation-invariant patterns. We train CNNs to classify attention weights into the classes of a given taxonomy. Our architecture consists of two convolutional layers with $\relu$ activations, dropout, and max-pooling, followed by two fully connected layers for classification. We also apply a layer-wise max-pooling and average-pooling over attention heads to focus on the most salient token-to-token interactions. Therefore, the CNNs have as many input channels as there are layers in the model. See implementation details and hyperparameters in the Appendix.

\subsection{Evaluation Metric: Minimum F1 Score} \label{sec:eval_metrics}

The CNN's test performance reflects how well the taxonomy aligns with the attention weights. For example, the second and fourth samples in Figure~\ref{fig:example_attention_weights} display similar diagonal patterns in the lower layers, leading the CNN to assign them the same class. A good taxonomy should group such samples together based on their shared patterns. Doing so reduces misclassifications and improves the CNN's test accuracy.

For each taxonomy, we randomly sample 4,000 training and 2,000 evaluation attention weights per class. This ensures balanced datasets, allowing us to assess the distinctive patterns of each memorization type regardless of its frequency. For each taxonomy, we train 8 CNNs with different hyperparameters (detailed in the Appendix). We also use 3 model sizes (Pythia 12B, 6.9B, and 2.8B) and evaluate the CNNs at 3 different steps (epochs 1, 2, and 3). We deliberately use limited data and training time to focus on taxonomies that are sufficiently salient in the attention weights to be learned quickly by the CNNs. This results in $2{,}000 \times 8 \times 3 \times 3 = 144{,}000$ test predictions per class and per taxonomy. We compute the precision, recall, and $F_1$ score for each class. To favor taxonomies that account for all forms of memorization and penalize those with one low-performing class, we focus on the minimum $F_1$ score across all classes and use it as our main evaluation metric.

\subsection{Parametrization of taxonomies} \label{sec:taxonomy_parametrization}

\input{figures/taxonomy_parametrization/taxonomy_prametrization}

To build a comprehensive benchmark of taxonomies, we developed a parametrization that allows exploration of a wide range of possible taxonomies. We model taxonomies as decision trees. In a taxonomy well-aligned with attention weights, each node should isolate samples that rely on meaningfully distinct memorization mechanisms. Based on prior work, we defined two families of nodes likely to influence the nature of memorization. The \textit{duplication-based} nodes, defined in the first frame of Figure~\ref{fig:taxonomy_parametrization}, separate samples using a duplication threshold $\hdelta$, as duplication is known to influence memorization \citep{carlini_quantifying_2023}. The \textit{completion-based} nodes, defined in the second frame, capture samples where most suffix tokens can be predicted from the prefix. The \textit{Reconstruct} node matches the definition from \citet{prashanth_recite_2024}. \textit{Guess}[$\hlambda, \hgamma$] expands on it using ROUGE-based conditions to include more samples. Finally, we added a \textit{Code} node, as the strict syntax of code strongly constrains the suffix. We also define rules to construct reasonable trees from these nodes and ensure that each class has a simple explanation. These rules are detailed in the Appendix and lead to the families presented in the last frame of Figure~\ref{fig:taxonomy_parametrization}. 

%% file: figures/taxonomy_parametrization/taxonomy_prametrization.tex
\begin{figure}[t!]
    \centering
    \includegraphics[width=\columnwidth]{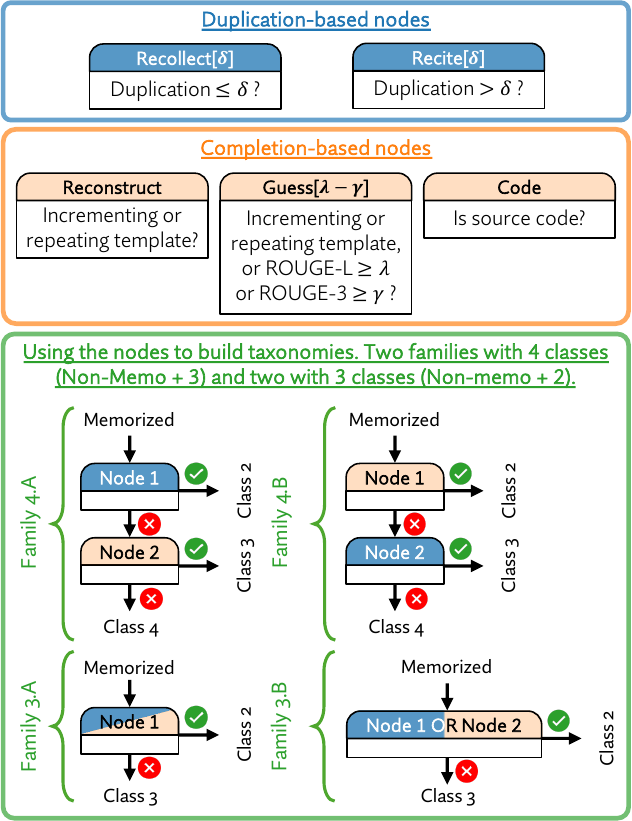}
    \caption{We parametrize taxonomies as decision trees with two types of nodes. We omit the \textit{Non-Memorized} node at the root of each taxonomy, because memorized samples are always defined as 32-extractable sequences.}
    \label{fig:taxonomy_parametrization}
\end{figure}

%% file: sections/04_empirical_results.tex
\section{Taxonomy Benchmark: Results} \label{sec:empirical_results}

\input{figures/confusion_matrices/example_confusion}

Our main empirical results are presented in Table~\ref{tab:taxonomy_results} and Figure~\ref{fig:example_confusion_matrix}. We evaluate the 54 possible taxonomies with $\delta \in \{5, 50, 1000\}$ and $\lambda = \gamma = 0.5$. Taxonomies are denoted by their list of nodes, with $\hdelta, \hlambda, \hgamma$ in brackets, and \textit{Others} to refer to the remaining samples. For example, \citet{prashanth_recite_2024}'s taxonomy is \textit{Non-Memo, Recite[5], Reconstruct, Others} ; ours is \textit{Non-Memo, Guess[0.5-0.5], Others}.

Table~\ref{tab:taxonomy_results} presents the list of possible taxonomies ranked by descending performance, grouped by number of classes. For brevity, we show only the most relevant taxonomies; full results are available in Table~\ref{tab:taxonomy_results_full} in the Appendix. The taxonomy of \citet{prashanth_recite_2024} performs poorly, with a minimum $F_1$ of 64.7\%, well below the best 4-classes taxonomy, which achieves 72.8\%. In contrast, the best 3-classes taxonomy outperform all others by a clear margin. Its confusion matrix shows very few misclassifications, indicating that its classes are well aligned with the attention weights.

To account for the increased difficulty of 4-classes classification, we normalize the $F_1$ score between a random predictor ($F_1^\text{rand} = 25\%$ or $33.3\%$) and a perfect one ($F_1^\text{max} = 100\%$) using: $F_1^\text{norm} = (F_1 - F_1^\text{rand}) / (F_1^\text{max} - F_1^\text{rand})$. After normalization, the best 3-classes taxonomy reaches $F_1^\text{norm} = 83.6\%$, compared to 63.7\% for the best 4-classes taxonomy. This supports the findings in Figure~\ref{fig:example_confusion_matrix}: the 4-classes taxonomy exhibits substantial misclassification, whereas the 3-classes taxonomy yields much cleaner separation. We therefore recommend the best 3-classes taxonomy, which we adopt as the data-driven taxonomy proposed in this paper: \textit{Non-Memo, Guess[0.5-0.5], Others}.

\subsection{The illusion of Few-shot Memorization}

We observe that \textit{Others} samples in \citet{prashanth_recite_2024}'s taxonomy are often misclassified, with a $F_1$ score of 64.7\%. These samples correspond to few-shot memorization (called \textit{Recollect}  in their work): they are supposedly memorized without being highly duplicated or without following a template. The numerous misclassification for this class demonstrate that it does not correspond to a distinct form of memorization. This aligns with the findings of \citet{huang_demystifying_2024}, which show that most samples believed to be few-shot memorized are either approximately duplicated in the dataset or follow templates not covered by the definition of \textit{Reconstruct}. It also supports the observation that random canaries must be duplicated at least a few dozen times to be memorized \citep{meeus_copyright_2024}.

Moreover, the two highest-ranking taxonomies in Table~\ref{tab:taxonomy_results} do not rely on duplication and outperform all others by a clear margin. This indicates that duplication does not trigger a distinct memorization mechanism, and there is no meaningful difference between the attention weights of a random canary and those of a software license duplicated 50 or 1000 times. While it is well established that duplication facilitates memorization \citep{carlini_quantifying_2023}, our experiments demonstrate that duplication is a necessary condition for verbatim memorization, but a high duplication rate does not qualitatively alter the nature of memorization.

\input{figures/tax_perf/tax_perf_small}

\subsection{Impact of ROUGE parameters}

We use $\hlambda = \hgamma = 0.5$ to define the \textit{Guess} class in our benchmark. This choice is intuitive, as it implies that half of the suffix tokens are constrained by the prefix. Since the highest-ranking taxonomy includes a \textit{Guess} node, we investigated whether its performance could be further improved by optimizing $\hlambda$ and $\hgamma$. To that end, we evaluated the taxonomy \textit{Non-Memo, Guess[$\hlambda$-$\hgamma$], Others} for $\hlambda = \hgamma \in \{0.1, 0.2, \dots, 0.9\}$. We also tested $\hlambda = 1$ with $\hgamma \in \{0.1, 0.2, \dots, 0.9\}$ (disabling the ROUGE-L condition), and vice versa. We found that optimizing $\hlambda$ and $\hgamma$ yields negligible improvements, increasing the minimum $F_1$ score by only 0.2\% (see Table~\ref{tab:taxonomy_results_rouge} in the Appendix). We therefore recommend using the most intuitive setting: $\hlambda = \hgamma = 0.5$.

\subsection{Impact of model size}

Our results are averaged over three model sizes: Pythia 12B, 6.9B, and 2.8B. We also evaluated the taxonomies on each size separately. We found that our highest-ranking taxonomy also ranks highest for each size, confirming that its classes accurately capture distinct attention mechanisms that persist across model scales. See Tables~\ref{tab:taxonomy_results_12b}-\ref{tab:taxonomy_results_2-8b} in the Appendix.

%% file: figures/confusion_matrices/example_confusion.tex
\begin{figure*}[t!]
    \centering
    \includegraphics[width=\textwidth]{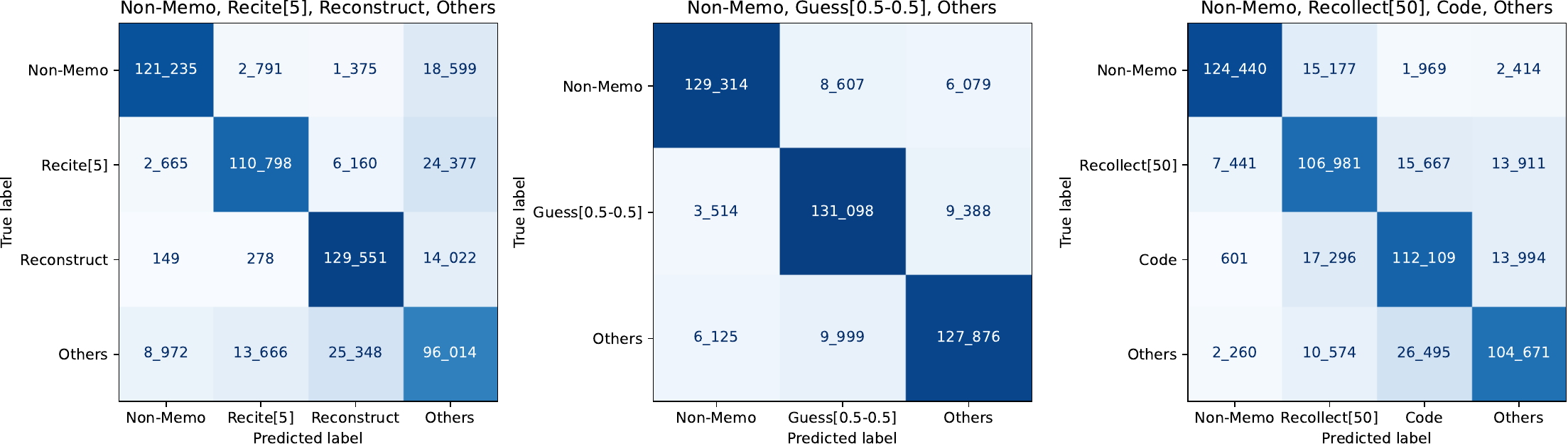}
    \caption{Confusion matrix for three taxonomies: \citet{prashanth_recite_2024} (left), ours (middle), and the best 4-classes taxonomy (right, see Table \ref{tab:taxonomy_results}). Datasets are balanced, with $144{,}000$ attention weights in each class.}
    \label{fig:example_confusion_matrix}
\end{figure*}

%% file: figures/tax_perf/tax_perf_small.tex
\begin{table}[!t]
    \centering
    \small
    \begin{tabular}{@{}l@{}cc@{}} 
        \toprule
        \textbf{Taxonomy name} & \textbf{Classes} & \textbf{Min $F_1$} \\
        \midrule
        Non-Memo, Recollect[50], Code, Others & 4 & 72.8 \\
        \textit{10 lines with $64.7\% < F_1 < 72.8\%$ omitted} & 4 & | \\
        $\blacklozenge$ \textbf{Non-Memo, Recite[5], Reconstruct, Others} & \textbf{4} & \textbf{64.7} \\
        \textit{15 lines with $F_1 < 64.7\%$ omitted} & 4 & | \\
        \midrule
        $\bigstar$ \textbf{Non-Memo, Guess[0.5-0.5], Others} & \textbf{3} & \textbf{89.0} \\
        Non-Memo, Reconstruct, Others & 3 & 87.7 \\
        \textit{25 lines with $F_1 \le 83.8\%$ omitted} & 3 & | \\
        \bottomrule
    \end{tabular}
    \caption{Taxonomy benchmark: Minimum $F_1$ across all categories for selected taxonomies. $\blacklozenge$ denotes \citet{prashanth_recite_2024}'s taxonomy. We adopt as our taxonomy the highest-ranking one, denoted by $\bigstar$. See full table in the Appendix.}
    \label{tab:taxonomy_results}
\end{table}

%% file: sections/05_localizing_memorization.tex
\section{Localizing memorization} \label{sec:localizing_memorization}

The optimal taxonomy derived from our benchmark is \textit{Non-Memo, Guess[0.5-0.5], Others}. However, "Others" is not an intuitive label for samples that are memorized without being guessed. For simplicity, we now refer to these classes as \textit{Non-Memo, Guess, Recall}, as in the Introduction. To demonstrate the benefits of studying \textit{Guess} and \textit{Recall} as distinct forms of memorization, we develop a custom interpretability technique to analyze the CNNs trained under this taxonomy and examine the regions of the attention weights that play a significant role in each form of memorization.

\subsection{Methodology} \label{sec:localizing_methodology}

The CNNs have one input channel per LLM layer, leading to more numerous and more heterogeneous channels than typical computer vision settings (up to 36 channels capturing diverse patterns). As a result, standard explainability methods for CNNs, such as GradCAM \cite{selvaraju_grad-cam_2017} lose their explainability capabilities under these conditions.

We therefore developed a custom interpretability technique. Consider a taxonomy with classes $\mathcal{U}_1, \mathcal{U}_2, \cdots, \mathcal{U}_N$ and a CNN $f$ trained on this taxonomy. We aim to compute matrices $\Delta_l[t_0] \in \mathbb{R}^{64 \times 64}$ for each class $t_0$ and layer $l$, representing the regions of the attention weights that contribute specifically to classifying samples in $\mathcal{U}_{t_0}$.

\paragraph{Step 1: Guided Backpropagation.}

Let $s \in \mathcal{U}_{t_0}$ be a sample with  ground truth label $\mathcal{U}_{t_0}$. Let $\mA^h_l[s]$ denote the attention weight at head $h$ and layer $l$. For each possible class $\mathcal{U}_{t_1}$, we apply Guided Backpropagation to sample $s$ with respect to that class \citep{springenberg_striving_2015}. It is similar to computing the gradient of the logit for $\mathcal{U}_{t_1}$ with respect to $s$ using a backward pass, except that negative gradients are clipped to zero at each layer. This gives us $\mB^h_l[s \rightarrow t_1] \in \mathbb{R}^{64 \times 64}$, which identifies the \mbox{coordinates} of the input that \textit{could} contribute positively to class~$\mathcal{U}_{t_1}$. Note that $\mB^h_l[s \rightarrow t_1]$ can have positive values at positions where $\mA^h_l[s]$ is zero.

\paragraph{Step 2: Discriminative classification} 

We then compute $\mC^h_l[s] \in \mathbb{R}^{64 \times 64}$, which captures the positions of attention weights that can contribute positively to the correct class of $s$ but not to the others. The goal is to identify \textit{discriminative} regions in the attention weights. For instance, the diagonal of $\mA^h_l[s]$ is always highly activated, since each token attends to itself; however, because this is not a discriminative feature, we aim to disregard it.

\begin{equation*}
\mC^h_l[s] = \mB^h_l[s \rightarrow t_0] - \frac{1}{N-1}\sum_{t_1 \neq t_0} \mB^h_l[s \rightarrow t_1]
\end{equation*}

\paragraph{Step 3: Clipping and normalization}

Then, we compute $\mD^h_l[s] \in \mathbb{R}^{64 \times 64}$ by clipping negatives values in $\mC^h_l[s]$ to focus only on positive influence, and normalize by the maximal value across all heads $h$, layers $l$, and positions $i,j \in \llbracket 0, 63 \rrbracket$. Note that the second $\max$ is element-wise.

\begin{equation*}
\mD^h_l[s] = \frac{1}{\max\limits_{l, h, i, j} \ \left[\mC^h_l[s]\right]_{i,j}} \times \max (\mC^h_l[s], 0)
\end{equation*}

\paragraph{Step 4: Activations and Averaging}

Finally, we multiply by the activations $\mA^h_l[s]$ to identify the positions that \textit{actively} influence class $\mathcal{U}_{t_0}$ for input~$s$. We apply layer-wise max-pooling over heads to focus on the most salient activations. Then, we average over all samples in $\mathcal{U}_{t_0}$ and all CNNs trained under this taxonomy. This yields $\Delta_l[t_0] \in \mathbb{R}^{64 \times 64}$, indicating the regions of the attention weights that \textit{consistently} influence that class. For simplicity, we omit the average over the CNNs in the notation.

\begin{equation*}
\Delta_l[t_0] = \frac{1}{|\mathcal{U}_{t_0}|} \sum_{s \in \mathcal{U}_{t_0}} 
\max\limits_{h} \left[ \mD^h_l[s] \times \mA^h_l[s] \right]
\end{equation*}

\subsection{Empirical Results} \label{sec:localizing_results}

\input{figures/localizing_results/localizing_results}

\paragraph{Syntactic, low-layer connections for \textit{Guess}}

Our main results are presented in Figure~\ref{fig:localizing_results}. We observe that the lower layers of the LLM contribute significantly to memorizing \textit{Guess} samples. The mean value of $\Delta_l[\verb|Guess|]$ is high in these layers, peaking at layer 6. At that layer, $\Delta_6[\verb|Guess|]$ displays typical diagonal patterns in the first half of the matrix (key token index $\le 31$). Layers 7–9 show similar behavior (see Figure~\ref{fig:full_delta_matrix_7_12} in the Appendix). As in the example from Figure~\ref{fig:example_attention_weights}, these diagonal segments reflect direct causal links between the prefix and the suffix, such as repetitions. 

In light of recent studies highlighting the role of higher layers, such as \citet{menta_analyzing_2025} and \citet{dentan_predicting_2025}, the significant contribution of lower layers to memorization was unexpected. We interpret this by noting that the \textit{Guess} class primarily captures low-level syntactic dependencies that do not require complex token interactions and can emerge in the earliest layers of the LLM. Similarly, \citet{stoehr_localizing_2024} observed that some lower attention heads contribute substantially to memorization. Upon analyzing their dataset, we found that 54\% of the samples are code snippets. As for the \textit{Guess} class, we interpret their findings as evidence of low-level dependencies between the prefix and the suffix due to syntactic regularity of code snippets. Conversely, as we will show, other forms of memorization rely more heavily on higher layers.

\paragraph{Short-range, high-layer connections for \textit{Recall}}

The mean value of $\Delta_l[\verb|Recall|]$ is particularly high in the higher layers of the model, peaking at the final layer. At that layer, $\Delta_{36}[\verb|Recall|]$ exhibits strong activation just \textit{below} the main diagonal. Layers 31–35 show similar behavior (see Figure~\ref{fig:full_delta_matrix_31_36} in the Appendix). The fact that \textit{Recall} relies on attention weights just \textit{below} the diagonal suggests that the model uses the few preceding tokens to complete each token. This corroborates the findings of \citet{huang_demystifying_2024}, who show that only a few tokens from a memorized sample are encoded by \textit{triggers} in the prefix (see their Figure 4). They explain that the remaining tokens are inferred using language modeling capabilities. Our results suggest that these capabilities are localized in the activations below the diagonal in $\Delta_l[\verb|Recall|]$ for $l \ge 31$. This indicates that the model relies on short-range connections in the final layers, which are highly correlated with the output, to fill in the gaps between the tokens encoded by the \textit{triggers}.

Importantly, our results suggest that \textit{Non-Memo} relies on different layers than \textit{Recall}. We observe that the mean value of $\Delta_l[\verb|Non-Memo|]$ is significantly higher in the intermediate layers of the LLM. This indicates that these attention blocks play a major role in the model’s general-purpose capabilities but contribute little to memorizing \textit{Recall} samples. This observation helps explain why \citet{menta_analyzing_2025} were able to reduce memorization while preserving overall performance by deactivating the final attention blocks of the model. As we have shown, these upper layers, which are highly correlated with the output, are crucial to fill in the gaps between memorized tokens for \textit{Recall}, but they appear to be less essential for the general-purpose abilities involved in decoding \textit{Non-Memo} samples.

%% file: figures/localizing_results/localizing_results.tex
\begin{figure}[t!]
    \centering
    \begin{subfigure}[b]{\columnwidth}
        \centering
        \includegraphics[width=\textwidth]{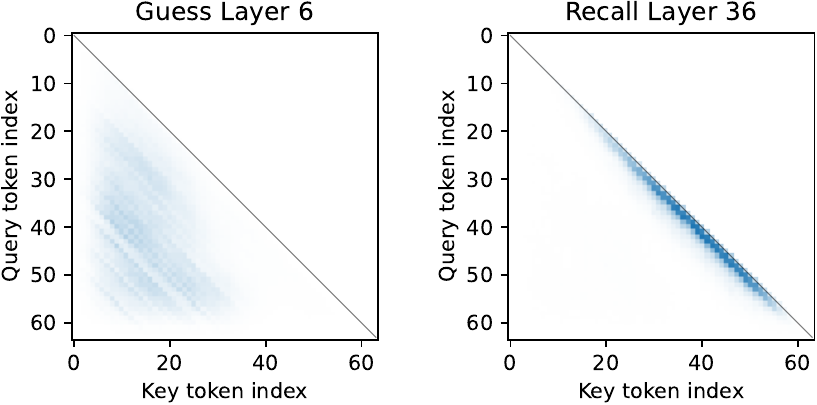}
        \caption{$\Delta_l[t_0]$ matrices at selected layers}
        \label{fig:example_delta_matrix}
    \end{subfigure}
    \begin{subfigure}[b]{\columnwidth}
        \centering
        \includegraphics[width=\textwidth, trim=0 0 0 -6pt, clip]{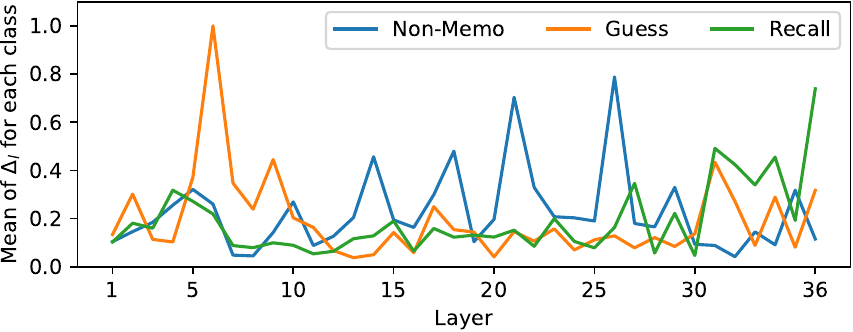}
        \caption{Mean value of $\Delta_l[t_0]$ across layers}
        \label{fig:layer_analysis}
    \end{subfigure}
    \caption{We compute $\Delta_l[t_0]$ for every class $t_0$ and layer $l$ using the CNNs trained on Pythia 12B attention weight under our optimal taxonomy: \textit{Non-Memorized, Guess, Recall. }Figure~\ref{fig:example_delta_matrix} shows $\Delta_l[t_0]$ for selected layers and classes (see all layers in the Appendix). Figure~\ref{fig:layer_analysis} presents the decisiveness of each layer for classification under this taxonomy.}
    \label{fig:localizing_results}
\end{figure}

%% file: sections/06_concluding_remarks.tex
\section{Limitations and future work} \label{sec:limitations}

\paragraph{Datasets and models}

A primary limitation of our work is that all experiments were conducted on models from the same family: Pythia 12B, 6.9B, and 2.8B \citep{biderman_pythia_2023}, using their training dataset, the Pile \citep{gao_pile_2020}. This choice was driven by the necessity of having access to the full training data of the analyzed models, which is unavailable for most open-source models. As a result, memorization research typically focuses on either GPT-NeoX \citep{black_gpt-neo_2021} or Pythia. We selected Pythia because it is used in the existing taxonomy we compare against \citep{prashanth_recite_2024}, but evaluating our approach on other models remains a promising direction for future work. To partially address this limitation, we performed experiments across multiple Pythia model sizes (see Section \textit{Impact of model size}). The consistent results observed across scales provide an incomplete but encouraging indication of the generality of our findings.

\paragraph{A focus on attention blocks}

Our approach focuses exclusively on the model's attention weights, disregarding the role of feed-forward blocks. This choice is motivated by two factors. First, the role of feed-forward layers has already been investigated in prior work, notably \citep{huang_demystifying_2024}. Second, we chose to concentrate on attention blocks to analyze the causal links between the prefix and the suffix that are essential for verbatim memorization. Nonetheless, future work could explore an evaluation of taxonomies that incorporates both attention and feed-forward blocks.

\paragraph{An indirect localization}

Finally, the explainability method we developed allows us to localize memorization indirectly by analyzing CNNs trained on attention weights. However, it would be valuable to correlate our findings with direct observations, such as ablations or perturbations of the attention weights. Similar ablations have been performed by \citet{menta_analyzing_2025}, and applying them separately to each form of memorization would be an interesting direction for future work.

\section{Conclusion} \label{sec:conclusion}

We show that existing taxonomies proposed in the literature for memorization in Large Language Models do not align with the attention mechanisms underlying verbatim memorization. To address this gap, we introduce a systematic approach for exploring and evaluating a broad set of candidate taxonomies. Based on this approach, we propose a new data-driven taxonomy that significantly outperforms all others: \textit{Non-Memorized, Guess, Recall}. We then developed a custom method to localize the regions of the attention weights that are critical for each of these forms of memorization.

Our results corroborate and extend several recent findings in the memorization literature. We confirm the illusion of verbatim memorization by showing that duplication does not induce a distinct form of memorization. We demonstrate that a significant proportion of samples are \textit{guessed} by the model using syntactic dependencies and highlight the importance of lower layers for that mechanism. Finally, we show that the language modeling abilities involved in memorization differ from the model's general-purpose abilities. They reside in short-range connections in the latest layers of the model and  control the exact decoding of memorized tokens. These findings underscore the importance of studying each form of memorization separately in future research.

\section{Ethical Considerations} \label{sec:ethics}

While our work advances the understanding of memorization in LLMs, it will benefit privacy researchers more than attackers, for several reasons. First, our experiments are conducted on public datasets, which are of no interest to an attacker. Moreover, we focus solely on analyzing memorization without introducing new attack methods. On the contrary, our findings can help develop more effective mitigation strategies.

To ensure reproducibility and to support further research on LLM memorization, we release all scripts needed to reproduce our experiments. Implementation details are discussed in the Appendix.

\begin{links}
    \link{Code}{https://github.com/orailix/cnn-4-llm-memo}
\end{links}

\acknowledge{
This work received financial support from Crédit Agricole SA through the research chair ``Trustworthy and Responsible AI'' with École Polytechnique. This work was granted access to the HPC resources of IDRIS under the allocation 2023-AD011014843 made by GENCI. Finally, we thank Mohamed Dhouib and Mathis Le Bail discussions on early versions of this paper. 
}

%% file: sections/a_01_implementation_details.tex
\section{Implementation details} \label{app:implementation_details}

\paragraph{CNN architecture} Our CNN architecture consists of two convolutional layers with $\relu$ activations, dropout, and max-pooling, followed by two fully connected layers for classification. To reduce the number of input channels, we apply layer-wise head pooling: for instance, the attention maps from Pythia 12B (with 36 layers and 40 heads per layer, totaling 1440 channels) are reduced to 36 channels by retaining only the maximal value across head for each position of the attention weight. The CNNs are trained using a cross-entropy loss.

\paragraph{CNN hyperparameters} The hyperparameters used to train our CNNs are detailed in Table \ref{tab:hyperparameters_cnns}. To ensure a robust evaluation, for each taxonomy, we train 8 distinct CNNs with different hyperparameter and collect the predictions of the 8 CNNs for our benchmark. Depending on these parameters, our CNNs have between $155{,}415$ and $325{,}540$ trainable parameters. Precision, Recall, and $F_1$ score are computed from a single training run for each of the 8 hyperparameter combinations.

\input{sections/a_01b_table_hpar}

\paragraph{Computing infrastructure} Experiments were conducted on an HPC cluster node equipped with 2× AMD EPYC 7543 CPUs (32 cores, 64 threads each), 468 GB of RAM, and 8× NVIDIA A100 GPUs (80 GB each), running Red Hat Enterprise Linux 9.4. Relevant software versions: Python 3.11.0, \verb|accelerate|~1.1.1, \verb|datasets|~3.1.0, \verb|numpy|~1.26.4, \verb|pandas|~2.2.3, \verb|rouge_score|~0.1.2, \verb|torch|~2.5.1, \verb|transformers|~4.46.3.

\paragraph{Computational time} Our experiments amount to a total of $2{,}190$~single-GPU-hours on NVIDIA A100.

\paragraph{ROUGE score} In this paper, we use the ROUGE F1 score computed with $\verb|rouge-score|$ library from PyPI.

\paragraph{Source code link, license and intended use}

We release the Python source code, as well as the Bash and Slurm scripts needed to reproduce all experiments presented in this paper. The license of this code is included in its repository, as well as the documentation of the project. This source code is intended to be used for research only.

\begin{links}
    \link{Code}{https://github.com/orailix/cnn-4-llm-memo}
\end{links}

%% file: sections/a_01b_table_hpar.tex
\begin{table}[ht]
    \centering
    \small
    \begin{tabular}{l|c|c} %
        \toprule
        \textbf{Parameter} & \textbf{Type} & \textbf{Value(s)} \\
        \midrule
        Attention head pooling & Variable & $\{\max, \average\}$ \\
        Convolution: num features & Variable & $\{10, 16\}$ \\
        Convolution: kernel size & Variable & $\{6, 8\}$ \\
        \midrule
        Convolution: pooling type & Fixed & $\max$ \\
        Convolution: pooling size & Fixed & $2$ \\
        Feed-forward: num features & Fixed & $64$ \\
        Dropout & Fixed & $0.5$ \\
        Activation function & Fixed & $\relu$ \\
        Train: batch size & Fixed & $16$ \\
        Train: learning rate & Fixed & $0.001$ \\
        Train: weight decay & Fixed & $0.1$ \\
        Train: num epoch & Fixed & $3$ \\
        \bottomrule
    \end{tabular}
    \caption{Hyperparameters of the CNNs trained under our taxonomies. Three hyperparameters variable values with two possibilities, leading to a total of $2^3 = 8$ distinct CNNs trained under each taxonomy.}
    \label{tab:hyperparameters_cnns}
\end{table}

%% file: sections/a_02_details_parametrization.tex
\section{Details on our taxonomy parametrization} \label{app:taxonomy_parametrization}

\paragraph{Rules for building decision trees}

Based on the \textit{duplication-based} and \textit{completion-based} nodes introduced in the first two frames of Figure \ref{fig:taxonomy_parametrization}, we implement three rules that constrain the decision trees defining our taxonomies. These rules ensure that each taxonomy is easily interpretable and remains simple, as our goal is to capture salient characteristics of the attention weights while ignoring patterns that pertain only to highly specific types of memorization defined by overly complex taxonomies.

\begin{enumerate}
    \item Each tree should contain at most one completion-based and one duplication-based node, as these two node types capture distinct mechanisms. Including the same type twice in a single tree would be redundant.
    \item The "yes" branch of a node should lead to a leaf, not another decision node. This ensures that our taxonomies identify \textit{positive} memorization features through the \textit{presence} of patterns in the attention weights, enhancing the explainability of the CNN decisions.
    \item 3-classes taxonomies should be derived from 4-classes trees by merging classes. As a result, only "OR" statements are allowed in 3-classes taxonomies, not "AND" statements. This prevents the creation of overly specific classes that might overfit to attention patterns.
\end{enumerate}

\paragraph{Counting possible taxonomies} The rules defined above yield the four families of taxonomies shown in the third frame of Figure \ref{fig:taxonomy_parametrization}. There are 27 four-class taxonomies and 27 three-class taxonomies. To verify this, we fix $\hdelta = 5$ and count the number of possible trees.

\begin{itemize}
    \item There are $9$ taxonomies with $4$ classes, all of which are $\hdelta$-dependent.
    \begin{enumerate}
        \item Family 4.A: $2 \times 3 = 6$ trees.
        \item Family 4.B: $3 \times 1 = 3$ trees. For Node 2, the classes remain the same (up to ordering) regardless of the choice of duplication-based node.
    \end{enumerate}
    \item There are $11$ taxonomies with $3$ classes: three are $\hdelta$-invariant, and eight are $\hdelta$-dependent.
    \begin{enumerate}
        \item Family 3.A: $2 + 3 = 5$ trees. The three taxonomies defined using only completion-based nodes are $\hdelta$-invariant.
        \item Family 3.B: $2 \times 3 = 6$ trees, all of which are $\hdelta$-dependent.
    \end{enumerate}
\end{itemize}

We use three values for $\hdelta$: ${5, 50, 1000}$. Consequently, there are $3 \times 9 = 27$ four-class taxonomies and $3 \times 8 + 3 = 27$ three-class taxonomies.

%% file: sections/a_03_additional_figures.tex
\section{Additional Results} \label{app:additional_figures}

In addition to the results presented in the main paper, we provide the full evaluation benchmark for all taxonomies (see Table \ref{tab:taxonomy_results_full}), an ablation study on the impact of $\hlambda$ and $\hgamma$ (see Table \ref{tab:taxonomy_results_rouge}), the evaluation benchmark grouped by model size (12B, 6.9B, and 2.8B; see Tables \ref{tab:taxonomy_results_12b}, \ref{tab:taxonomy_results_6-9b}, and \ref{tab:taxonomy_results_2-8b}), the confusion matrices of CNNs trained under 4-classes taxonomies (see Figures \ref{fig:full_confusion_4_classes_part_1}, \ref{fig:full_confusion_4_classes_part_2} , and \ref{fig:full_confusion_4_classes_part_3}) and 3-classes taxonomies (see Figures \ref{fig:full_confusion_3_classes_part_1}, \ref{fig:full_confusion_3_classes_part_2}, and \ref{fig:full_confusion_3_classes_part_3}), and visualizations of $\Delta_l[t_0]$ for all classes of our optimal taxonomy across all layers (see Figures \ref{fig:full_delta_matrix_31_36}, \ref{fig:full_delta_matrix_25_30}, \ref{fig:full_delta_matrix_19_24}, \ref{fig:full_delta_matrix_13_18}, \ref{fig:full_delta_matrix_7_12}, and \ref{fig:full_delta_matrix_1_6}).

\clearpage
\input{figures/tax_perf/tax_perf_large}
\input{figures/tax_perf/tax_perf_rouge}
\input{figures/tax_perf/tax_perf_12b}
\input{figures/tax_perf/tax_perf_6-9b}
\input{figures/tax_perf/tax_perf_2-8b}
\input{figures/confusion_matrices/full_confusion_4_classes}
\input{figures/confusion_matrices/full_confusion_3_classes}
\input{figures/delta_matrices/full_delta}

%% file: figures/tax_perf/tax_perf_large.tex
\begin{table*}[!ht]
    \centering
    \small
    \begin{tabular}{@{}l@{}cccccccc@{}} 
        \toprule
        \textbf{Taxonomy name} & \textbf{Classes} & \textbf{Min $F_1$} & \textbf{Mean $F_1$} & \textbf{Min Prec} & \textbf{Mean Prec} & \textbf{Min Rec} & \textbf{Mean Rec} & \textbf{Mean Loss} \\
        \midrule
        Non-Memo, Recollect[50], Code, Others & 4 & \textbf{72.8} & 77.9 & \textbf{71.3} & 78.2 & 72.7 & 77.8 & 0.0450 \\
        Non-Memo, Recollect[5], Code, Others & 4 & \underline{72.0} & 78.4 & \underline{70.5} & 79.1 & 71.2 & 78.3 & 0.0442 \\
        Non-Memo, Recollect[5], Guess, Others & 4 & 71.5 & 79.8 & 65.7 & 80.6 & 68.4 & 79.7 & 0.0402 \\
        Non-Memo, Recite[5], Guess, Others & 4 & 70.3 & 79.2 & 63.3 & 80.6 & 70.0 & 78.9 & 0.0435 \\
        Non-Memo, Recollect[50], Guess, Others & 4 & 69.2 & 79.8 & 66.9 & 80.0 & 71.7 & 79.6 & 0.0402 \\
        Non-Memo, Guess, Recite[5], Others & 4 & 69.1 & 79.5 & 63.3 & 80.4 & \textbf{76.1} & 79.1 & 0.0416 \\
        Non-Memo, Recollect[5], Reconstruct, Others & 4 & 68.3 & \underline{80.6} & 61.2 & \underline{81.9} & \underline{75.4} & 80.1 & 0.0393 \\
        Non-Memo, Recollect[50], Reconstruct, Others & 4 & 68.3 & \textbf{81.6} & 63.8 & \textbf{82.3} & 73.4 & \textbf{81.3} & \textbf{0.0373} \\
        Non-Memo, Recite[50], Guess, Others & 4 & 65.7 & 77.8 & 62.5 & 78.4 & 69.1 & 77.6 & 0.0437 \\
        Non-Memo, Reconstruct, Recite[5], Others & 4 & 65.2 & 79.4 & 60.9 & 80.1 & 70.1 & 79.0 & 0.0412 \\
        Non-Memo, Recite[1k], Guess, Others & 4 & 64.8 & 77.0 & 62.9 & 77.3 & 65.8 & 76.9 & 0.0416 \\
        $\blacklozenge$ \textbf{Non-Memo, Recite[5], Reconstruct, Others} & 4 & 64.7 & 79.6 & 62.7 & 80.1 & 66.7 & 79.4 & 0.0414 \\
        Non-Memo, Guess, Recite[50], Others & 4 & 64.5 & 77.8 & 61.5 & 78.4 & 67.9 & 77.5 & 0.0427 \\
        Non-Memo, Recite[1k], Code, Others & 4 & 63.9 & 71.3 & 61.4 & 71.6 & 61.3 & 71.1 & 0.0502 \\
        Non-Memo, Code, Recite[5], Others & 4 & 63.7 & 72.4 & 58.8 & 73.7 & 58.3 & 72.1 & 0.0495 \\
        Non-Memo, Recollect[1k], Reconstruct, Others & 4 & 63.1 & \underline{80.2} & 62.7 & 80.2 & 63.6 & 80.2 & 0.0355 \\
        Non-Memo, Reconstruct, Recite[50], Others & 4 & 62.4 & 79.4 & 62.3 & 79.5 & 62.5 & 79.4 & 0.0406 \\
        Non-Memo, Recite[50], Reconstruct, Others & 4 & 61.8 & 79.0 & 62.6 & 79.2 & 61.0 & 79.1 & 0.0412 \\
        Non-Memo, Guess, Recite[1k], Others & 4 & 61.5 & 77.0 & 62.4 & 77.0 & 60.7 & 77.1 & 0.0404 \\
        Non-Memo, Recollect[1k], Guess, Others & 4 & 61.1 & 77.5 & 63.8 & 77.6 & 58.6 & 77.6 & 0.0402 \\
        Non-Memo, Recollect[1k], Code, Others & 4 & 59.7 & 72.3 & 63.6 & 72.6 & 56.3 & 72.3 & 0.0484 \\
        Non-Memo, Recite[1k], Reconstruct, Others & 4 & 59.7 & 78.0 & 63.0 & 77.9 & 56.7 & 78.5 & 0.0392 \\
        Non-Memo, Code, Recite[50], Others & 4 & 59.1 & 69.3 & 53.4 & 70.4 & 59.1 & 68.9 & 0.0512 \\
        Non-Memo, Reconstruct, Recite[1k], Others & 4 & 59.1 & 78.2 & 61.9 & 78.1 & 56.5 & 78.5 & \underline{0.0386} \\
        Non-Memo, Code, Recite[1k], Others & 4 & 55.4 & 68.0 & 60.0 & 68.8 & 49.8 & 68.3 & 0.0511 \\
        Non-Memo, Recite[5], Code, Others & 4 & 49.8 & 70.0 & 55.8 & 70.9 & 44.9 & 70.1 & 0.0504 \\
        Non-Memo, Recite[50], Code, Others & 4 & 49.7 & 69.9 & 57.0 & 70.6 & 44.0 & 70.2 & 0.0512 \\
        \midrule
        $\bigstar$ \textbf{Non-Memo, Guess, Others} & 3 & \textbf{89.0} & \underline{89.9} & \underline{87.6} & \underline{89.9} & \textbf{88.8} & \textbf{89.9} & \underline{0.0242} \\
        Non-Memo, Reconstruct, Others & 3 & \underline{87.7} & \textbf{90.7} & \textbf{89.9} & \textbf{90.9} & 83.4 & \textbf{90.8} & \textbf{0.0228} \\
        Non-Memo, Guess-or-Recollect[5], Others & 3 & 83.8 & 87.3 & 83.7 & 87.4 & 84.0 & 87.3 & 0.0278 \\
        Non-Memo, Reconstruct-or-Recollect[5], Others & 3 & 81.7 & 85.9 & 78.9 & 86.1 & \underline{84.7} & 85.8 & 0.0304 \\
        Non-Memo, Recite[5], Others & 3 & 80.8 & 85.1 & 77.7 & 85.4 & 84.1 & 85.0 & 0.0308 \\
        Non-Memo, Recollect[5], Others & 3 & 80.4 & 84.6 & 76.6 & 85.0 & 82.7 & 84.5 & 0.0314 \\
        Non-Memo, Guess-or-Recollect[50], Others & 3 & 80.3 & 85.6 & 82.7 & 85.7 & 78.0 & 85.7 & 0.0294 \\
        Non-Memo, Reconstruct-or-Recollect[50], Others & 3 & 78.0 & 84.2 & 77.6 & 84.4 & 78.5 & 84.2 & 0.0316 \\
        Non-Memo, Recite[5]-or-Reconstruct, Others & 3 & 77.7 & 82.8 & 72.6 & 83.4 & 78.4 & 82.6 & 0.0330 \\
        Non-Memo, Recollect[50], Others & 3 & 77.4 & 83.7 & 76.5 & 83.9 & 78.3 & 83.6 & 0.0323 \\
        Non-Memo, Recite[50], Others & 3 & 76.5 & 83.2 & 76.9 & 83.4 & 76.1 & 83.2 & 0.0325 \\
        Non-Memo, Code-or-Recollect[5], Others & 3 & 76.0 & 81.7 & 72.6 & 82.0 & 75.2 & 81.6 & 0.0325 \\
        Non-Memo, Code, Others & 3 & 75.6 & 81.3 & 72.1 & 82.0 & 70.5 & 81.3 & 0.0338 \\
        Non-Memo, Recite[5]-or-Code, Others & 3 & 73.9 & 79.2 & 65.9 & 81.0 & 68.0 & 79.0 & 0.0369 \\
        Non-Memo, Code-or-Recollect[50], Others & 3 & 73.6 & 80.5 & 71.0 & 80.7 & 74.6 & 80.4 & 0.0334 \\
        Non-Memo, Recite[50]-or-Reconstruct, Others & 3 & 73.4 & 80.6 & 70.0 & 81.0 & 77.1 & 80.4 & 0.0346 \\
        Non-Memo, Recite[1k]-or-Code, Others & 3 & 72.0 & 78.5 & 71.0 & 78.5 & 72.7 & 78.4 & 0.0361 \\
        Non-Memo, Recite[5]-or-Guess, Others & 3 & 71.2 & 77.6 & 63.6 & 79.2 & 68.6 & 77.2 & 0.0377 \\
        Non-Memo, Guess-or-Recollect[1k], Others & 3 & 69.9 & 80.5 & 74.7 & 80.9 & 65.2 & 80.7 & 0.0325 \\
        Non-Memo, Recite[1k], Others & 3 & 69.4 & 79.5 & 70.3 & 79.8 & 68.6 & 79.4 & 0.0339 \\
        Non-Memo, Recite[50]-or-Code, Others & 3 & 69.4 & 75.8 & 61.1 & 77.9 & 62.0 & 75.5 & 0.0387 \\
        Non-Memo, Recollect[1k], Others & 3 & 68.8 & 79.3 & 70.6 & 79.5 & 67.1 & 79.2 & 0.0340 \\
        Non-Memo, Recite[50]-or-Guess, Others & 3 & 68.6 & 75.6 & 61.4 & 77.2 & 64.6 & 75.2 & 0.0394 \\
        Non-Memo, Recite[1k]-or-Guess, Others & 3 & 68.2 & 76.3 & 65.7 & 76.8 & 62.8 & 76.3 & 0.0366 \\
        Non-Memo, Reconstruct-or-Recollect[1k], Others & 3 & 68.0 & 78.9 & 71.0 & 79.0 & 65.3 & 79.0 & 0.0339 \\
        Non-Memo, Recite[1k]-or-Reconstruct, Others & 3 & 66.7 & 75.9 & 62.3 & 76.5 & 66.1 & 75.7 & 0.0364 \\
        Non-Memo, Code-or-Recollect[1k], Others & 3 & 66.1 & 76.8 & 65.6 & 77.0 & 66.7 & 76.6 & 0.0360 \\
        \bottomrule
    \end{tabular}
    \caption{Evaluation results for different taxonomies. In the first column, $\blacklozenge$ denotes \citet{prashanth_recite_2024}, and $\bigstar$ denotes the highest-ranking taxonomy, which we adopted as our taxonomy. Both of them are described in Figure~\ref{fig:teaser}. We report the minimum (resp. mean) $F_1$ score across all categories within each taxonomy. We report similar values for the Precision (Prec) and the Recall (Rec). Finally, we report the mean evaluation loss of the CNNs, computed using the cross-entropy loss as during training.}
    \label{tab:taxonomy_results_full}
\end{table*}

%% file: figures/tax_perf/tax_perf_rouge.tex
\begin{table*}[!ht]
    \centering
    \small
    \begin{tabular}{@{}cc|ccccccc@{}} 
        \toprule
        \textbf{$\lambda$ (Rouge-L)} & \textbf{$\gamma$ (Rouge-3)} & \textbf{Min $F_1$} & \textbf{Mean $F_1$} & \textbf{Min Prec} & \textbf{Mean Prec} & \textbf{Min Rec} & \textbf{Mean Rec} & \textbf{Mean Loss} \\
        \midrule
            0.6 & 0.6 & \textbf{89.2} & \underline{90.4} & 87.3 & 90.5 & 87.1 & 90.4 & 0.0241 \\
            0.6 & $-$ & \underline{89.0} & 90.3 & 87.0 & 90.4 & 87.2 & 90.3 & 0.0241 \\
            $\bigstar$ \textbf{0.5} & $\bigstar$ \textbf{0.5} & 88.8 & 89.7 & 87.1 & 89.7 & \textbf{88.4} & 89.7 & 0.0245 \\
            0.5 & $-$ & 88.5 & 89.4 & 86.6 & 89.5 & \underline{88.1} & 89.4 & 0.0249 \\
            0.7 & $-$ & 88.4 & 90.3 & 86.7 & 90.5 & 85.3 & 90.4 & 0.0238 \\
            0.9 & 0.9 & 88.4 & 90.8 & 87.5 & \textbf{91.1} & 84.8 & \textbf{90.9} & \textbf{0.0230} \\
            0.7 & 0.7 & 88.3 & 90.3 & 86.9 & 90.5 & 85.5 & 90.3 & 0.0238 \\
            $-$ & 0.5 & 88.3 & 89.8 & 86.5 & 89.9 & 85.4 & 89.8 & 0.0251 \\
            0.9 & $-$ & 88.0 & \textbf{90.6} & \underline{87.7} & 90.8 & 84.4 & \underline{90.7} & \underline{0.0237} \\
            0.8 & 0.8 & 87.9 & 90.3 & 86.6 & \underline{90.6} & 84.9 & 90.4 & 0.0246 \\
            $-$ & 0.4 & 87.8 & 89.3 & 85.7 & 89.4 & 85.7 & 89.3 & 0.0253 \\
            $-$ & 0.6 & 87.8 & 89.8 & 86.5 & 90.0 & 84.6 & 89.8 & 0.0246 \\
            0.8 & $-$ & 87.7 & 90.1 & 87.2 & 90.3 & 84.0 & 90.1 & 0.0243 \\
            $-$ & 0.7 & 87.7 & 90.2 & 87.0 & 90.4 & 84.2 & 90.2 & 0.0243 \\
            $-$ & 0.3 & 87.6 & 88.8 & 85.4 & 88.8 & 86.9 & 88.7 & 0.0257 \\
            $-$ & 0.9 & 87.4 & \underline{90.4} & \textbf{88.1} & 90.5 & 83.4 & 90.4 & 0.0234 \\
            $-$ & 0.8 & 87.2 & 89.9 & 87.6 & 90.0 & 83.9 & 89.9 & 0.0239 \\
            $-$ & 0.2 & 86.9 & 88.6 & 84.7 & 88.7 & 87.0 & 88.6 & 0.0264 \\
            0.4 & $-$ & 86.5 & 88.4 & 85.9 & 88.4 & 87.1 & 88.4 & 0.0260 \\
            0.4 & 0.4 & 86.1 & 88.2 & 86.3 & 88.2 & 85.9 & 88.2 & 0.0261 \\
            0.3 & $-$ & 84.4 & 87.3 & 84.3 & 87.3 & 84.5 & 87.3 & 0.0268 \\
            $-$ & 0.1 & 84.1 & 87.2 & 83.7 & 87.3 & 84.4 & 87.2 & 0.0279 \\
            0.3 & 0.3 & 83.7 & 87.3 & 84.3 & 87.3 & 81.6 & 87.3 & 0.0269 \\
            0.2 & 0.2 & 77.3 & 83.8 & 77.9 & 84.2 & 73.5 & 83.8 & 0.0304 \\
            0.2 & $-$ & 77.1 & 83.5 & 78.2 & 83.7 & 74.3 & 83.5 & 0.0299 \\
            0.1 & 0.1 & 74.4 & 82.0 & 75.7 & 82.3 & 70.5 & 82.0 & 0.0324 \\
            0.1 & $-$ & 74.0 & 81.5 & 75.8 & 81.6 & 72.2 & 81.4 & 0.0326 \\
        \bottomrule
    \end{tabular}
    \caption{Evaluating the impact of $\hlambda$ and $\hgamma$ on the performance of the taxonomy that ranked highest in our benchmark: \textit{Non-Memo, Guess[$\hlambda$, $\hgamma$], Others} (see Table~\ref{tab:taxonomy_results_full}). We tested configurations with $\hlambda = \hgamma \in {0.1, 0.2, \dots, 0.9}$, as well as asymmetric settings: $\hlambda = 1$ (disabling the ROUGE-L condition, marked with $-$) with varying $\hgamma$, and conversely $\hgamma = 1$ with varying $\hlambda$. The value $\hlambda = \hgamma = 0.5$, marked with $\bigstar$, was selected for our final taxonomy as it achieved near-optimal performance while being intuitive, indicating that half of the suffix tokens are constrained by the prefix.}
    \label{tab:taxonomy_results_rouge}
\end{table*}

%% file: figures/tax_perf/tax_perf_12b.tex
\begin{table*}[!ht]
    \centering
    \small
    \begin{tabular}{@{}l@{}cccccccc@{}} 
        \toprule
        \textbf{Taxonomy name} & \textbf{Classes} & \textbf{Min $F_1$} & \textbf{Mean $F_1$} & \textbf{Min Prec} & \textbf{Mean Prec} & \textbf{Min Rec} & \textbf{Mean Rec} & \textbf{Mean Loss} \\
        \midrule
        Non-Memo, Recollect[5], Code, Others & 4 & \textbf{71.8} & 78.4 & \textbf{70.6} & 78.8 & 69.7 & 78.3 & 0.0447 \\
        Non-Memo, Recollect[50], Code, Others & 4 & \underline{71.0} & 77.7 & \underline{69.8} & 78.0 & 72.1 & 77.5 & 0.0453 \\
        Non-Memo, Recollect[5], Guess, Others & 4 & 70.1 & 79.9 & 65.5 & 80.6 & 69.2 & \underline{79.8} & 0.0399 \\
        Non-Memo, Guess, Recite[5], Others & 4 & 68.9 & 79.6 & 64.5 & 80.4 & \underline{74.0} & 79.2 & 0.0412 \\
        Non-Memo, Recite[5], Guess, Others & 4 & 68.7 & 78.4 & 62.8 & 79.6 & 70.2 & 78.1 & 0.0441 \\
        Non-Memo, Recollect[5], Reconstruct, Others & 4 & 67.8 & \textbf{80.5} & 60.9 & \textbf{81.8} & \textbf{76.6} & \textbf{80.0} & 0.0396 \\
        Non-Memo, Recollect[50], Guess, Others & 4 & 67.6 & 79.6 & 66.1 & 79.8 & 69.2 & 79.5 & 0.0402 \\
        Non-Memo, Recite[1k], Guess, Others & 4 & 66.1 & 78.2 & 67.3 & 78.4 & 65.0 & 78.2 & 0.0406 \\
        Non-Memo, Recollect[50], Reconstruct, Others & 4 & 65.6 & \underline{80.0} & 59.8 & \underline{81.1} & 72.7 & 79.5 & 0.0391 \\
        Non-Memo, Recite[1k], Code, Others & 4 & 65.2 & 72.2 & 62.8 & 72.3 & 63.9 & 72.2 & 0.0497 \\
        Non-Memo, Recite[50], Guess, Others & 4 & 64.8 & 77.1 & 63.1 & 77.6 & 66.6 & 77.1 & 0.0441 \\
        Non-Memo, Code, Recite[5], Others & 4 & 64.0 & 71.6 & 57.2 & 72.9 & 58.3 & 71.2 & 0.0501 \\
        $\blacklozenge$ \textbf{Non-Memo, Recite[5], Reconstruct, Others} & 4 & 63.4 & 79.0 & 60.7 & 79.8 & 66.4 & 78.8 & 0.0425 \\
        Non-Memo, Guess, Recite[50], Others & 4 & 63.1 & 77.4 & 60.9 & 77.7 & 65.5 & 77.1 & 0.0424 \\
        Non-Memo, Guess, Recite[1k], Others & 4 & 63.0 & 77.8 & 62.6 & 77.9 & 63.5 & 77.7 & 0.0401 \\
        Non-Memo, Reconstruct, Recite[5], Others & 4 & 62.7 & 78.3 & 59.3 & 79.1 & 66.5 & 78.0 & 0.0422 \\
        Non-Memo, Recollect[1k], Guess, Others & 4 & 62.2 & 77.8 & 63.1 & 78.1 & 61.4 & 77.7 & 0.0395 \\
        Non-Memo, Reconstruct, Recite[50], Others & 4 & 61.2 & 78.9 & 61.5 & 79.4 & 60.9 & 78.9 & 0.0416 \\
        Non-Memo, Reconstruct, Recite[1k], Others & 4 & 60.0 & 78.9 & 65.4 & 78.8 & 55.5 & 79.4 & \underline{0.0378} \\
        Non-Memo, Recollect[1k], Reconstruct, Others & 4 & 60.0 & 79.1 & 62.0 & 78.9 & 58.1 & 79.3 & \textbf{0.0356} \\
        Non-Memo, Recite[1k], Reconstruct, Others & 4 & 59.7 & 77.7 & 63.3 & 77.6 & 56.4 & 78.3 & 0.0385 \\
        Non-Memo, Recite[50], Reconstruct, Others & 4 & 59.3 & 77.9 & 61.8 & 78.1 & 57.0 & 78.2 & 0.0417 \\
        Non-Memo, Recollect[1k], Code, Others & 4 & 58.6 & 71.7 & 62.6 & 72.0 & 55.1 & 71.7 & 0.0488 \\
        Non-Memo, Code, Recite[50], Others & 4 & 55.4 & 69.1 & 52.7 & 69.8 & 58.4 & 68.6 & 0.0510 \\
        Non-Memo, Code, Recite[1k], Others & 4 & 54.6 & 68.3 & 60.3 & 68.8 & 49.7 & 68.5 & 0.0508 \\
        Non-Memo, Recite[50], Code, Others & 4 & 53.4 & 70.5 & 60.8 & 71.3 & 47.5 & 70.7 & 0.0510 \\
        Non-Memo, Recite[5], Code, Others & 4 & 49.6 & 70.1 & 55.1 & 71.0 & 45.1 & 70.2 & 0.0503 \\
        \midrule
        $\bigstar$ \textbf{Non-Memo, Guess, Others} & 3 & \textbf{89.2} & \underline{90.0} & \underline{86.9} & \underline{90.1} & \textbf{88.4} & \underline{90.0} & \underline{ 0.0244} \\
        Non-Memo, Reconstruct, Others & 3 & \underline{87.5} & \textbf{90.2} & \textbf{88.7} & \textbf{90.4} & 83.6 & \textbf{90.3} & \textbf{0.0238} \\
        Non-Memo, Guess-or-Recollect[5], Others & 3 & 84.0 & 87.4 & 82.0 & 87.6 & \underline{86.0} & 87.4 & 0.0274 \\
        Non-Memo, Reconstruct-or-Recollect[5], Others & 3 & 80.8 & 85.5 & 78.2 & 85.8 & 83.5 & 85.4 & 0.0305 \\
        Non-Memo, Recite[5], Others & 3 & 80.0 & 84.8 & 77.7 & 85.1 & 82.4 & 84.7 & 0.0311 \\
        Non-Memo, Recollect[5], Others & 3 & 78.5 & 83.4 & 74.3 & 83.9 & 81.4 & 83.2 & 0.0323 \\
        Non-Memo, Guess-or-Recollect[50], Others & 3 & 77.9 & 84.1 & 81.3 & 84.2 & 74.9 & 84.2 & 0.0307 \\
        Non-Memo, Recite[5]-or-Reconstruct, Others & 3 & 76.8 & 82.4 & 73.1 & 82.8 & 79.9 & 82.3 & 0.0328 \\
        Non-Memo, Code-or-Recollect[5], Others & 3 & 76.4 & 82.1 & 74.0 & 82.2 & 76.4 & 82.1 & 0.0320 \\
        Non-Memo, Code, Others & 3 & 76.2 & 81.4 & 74.3 & 81.5 & 73.6 & 81.3 & 0.0335 \\
        Non-Memo, Recollect[50], Others & 3 & 76.1 & 83.0 & 75.0 & 83.3 & 77.1 & 82.9 & 0.0329 \\
        Non-Memo, Reconstruct-or-Recollect[50], Others & 3 & 75.7 & 83.1 & 76.4 & 83.3 & 75.0 & 83.0 & 0.0319 \\
        Non-Memo, Recite[50], Others & 3 & 74.4 & 82.0 & 74.3 & 82.2 & 74.6 & 81.9 & 0.0338 \\
        Non-Memo, Recite[1k]-or-Code, Others & 3 & 74.2 & 79.7 & 72.6 & 79.8 & 73.2 & 79.7 & 0.0346 \\
        Non-Memo, Recite[5]-or-Code, Others & 3 & 73.9 & 79.2 & 64.8 & 81.5 & 66.8 & 78.9 & 0.0371 \\
        Non-Memo, Code-or-Recollect[50], Others & 3 & 72.7 & 80.2 & 71.1 & 80.3 & 74.5 & 80.0 & 0.0332 \\
        Non-Memo, Recite[50]-or-Reconstruct, Others & 3 & 71.2 & 79.8 & 69.9 & 80.1 & 72.5 & 79.6 & 0.0350 \\
        Non-Memo, Guess-or-Recollect[1k], Others & 3 & 70.3 & 80.8 & 75.3 & 81.2 & 65.6 & 81.0 & 0.0319 \\
        Non-Memo, Recite[5]-or-Guess, Others & 3 & 70.2 & 77.7 & 65.3 & 78.5 & 73.5 & 77.3 & 0.0367 \\
        Non-Memo, Reconstruct-or-Recollect[1k], Others & 3 & 68.5 & 78.9 & 70.0 & 78.9 & 67.1 & 79.0 & 0.0336 \\
        Non-Memo, Recollect[1k], Others & 3 & 68.5 & 79.0 & 69.9 & 79.3 & 67.1 & 79.0 & 0.0336 \\
        Non-Memo, Recite[1k], Others & 3 & 68.2 & 79.4 & 72.2 & 80.0 & 64.6 & 79.5 & 0.0340 \\
        Non-Memo, Recite[1k]-or-Guess, Others & 3 & 67.6 & 75.8 & 66.2 & 76.1 & 63.3 & 75.9 & 0.0360 \\
        Non-Memo, Recite[50]-or-Code, Others & 3 & 67.5 & 74.7 & 60.4 & 76.3 & 62.1 & 74.4 & 0.0387 \\
        Non-Memo, Recite[50]-or-Guess, Others & 3 & 67.3 & 74.4 & 60.2 & 75.9 & 61.3 & 74.1 & 0.0391 \\
        Non-Memo, Recite[1k]-or-Reconstruct, Others & 3 & 67.0 & 75.8 & 62.2 & 76.5 & 64.5 & 75.6 & 0.0359 \\
        Non-Memo, Code-or-Recollect[1k], Others & 3 & 62.3 & 75.4 & 65.9 & 75.7 & 59.0 & 75.5 & 0.0356 \\
        \bottomrule
    \end{tabular}
    \caption{Same experiments as in Table~\ref{tab:taxonomy_results_full}, restricted to the Pythia 12B model. Our data-driven taxonomy, marked with $\bigstar$, also achieves the highest performance when this model is evaluated in isolation. Similar results for Pythia 6.9B and Pythia 2.8B are provided in Tables \ref{tab:taxonomy_results_6-9b} and \ref{tab:taxonomy_results_2-8b}.}
    \label{tab:taxonomy_results_12b}
\end{table*}

%% file: figures/tax_perf/tax_perf_6-9b.tex
\begin{table*}[!ht]
    \centering
    \small
    \begin{tabular}{@{}l@{}cccccccc@{}} 
        \toprule
        \textbf{Taxonomy name} & \textbf{Classes} & \textbf{Min $F_1$} & \textbf{Mean $F_1$} & \textbf{Min Prec} & \textbf{Mean Prec} & \textbf{Min Rec} & \textbf{Mean Rec} & \textbf{Mean Loss} \\
        \midrule
        Non-Memo, Recite[5], Guess, Others & 4 & \underline{\textbf{71.5}} & 79.8 & 64.9 & 81.1 & 70.4 & 79.6 & 0.0437 \\
        Non-Memo, Recollect[5], Code, Others & 4 & \underline{\textbf{71.5}} & 77.9 & \textbf{69.5} & 78.6 & 69.3 & 77.9 & 0.0447 \\
        Non-Memo, Recollect[50], Code, Others & 4 & 70.8 & 76.8 & \underline{69.2} & 77.1 & 70.6 & 76.6 & 0.0464 \\
        Non-Memo, Recollect[5], Guess, Others & 4 & 70.7 & 79.3 & 65.7 & 79.8 & 68.2 & 79.2 & 0.0404 \\
        Non-Memo, Guess, Recite[5], Others & 4 & 68.1 & 79.1 & 62.4 & 80.0 & \textbf{75.1} & 78.7 & 0.0415 \\
        Non-Memo, Recollect[50], Reconstruct, Others & 4 & 67.8 & \textbf{81.2} & 63.7 & \textbf{81.9} & 72.4 & \textbf{80.9} & 0.0370 \\
        Non-Memo, Recollect[50], Guess, Others & 4 & 67.4 & 79.4 & 67.8 & 79.5 & 67.0 & 79.5 & 0.0402 \\
        Non-Memo, Recollect[5], Reconstruct, Others & 4 & 67.3 & 79.4 & 59.2 & \underline{81.2} & \underline{72.5} & 78.8 & 0.0392 \\
        Non-Memo, Reconstruct, Recite[5], Others & 4 & 66.0 & \underline{79.8} & 62.4 & 80.3 & 70.1 & 79.5 & 0.0406 \\
        Non-Memo, Recite[50], Guess, Others & 4 & 65.5 & 77.6 & 62.6 & 78.0 & 68.7 & 77.4 & 0.0437 \\
        $\blacklozenge$ \textbf{Non-Memo, Recite[5], Reconstruct, Others} & 4 & 64.4 & 79.4 & 63.7 & 79.7 & 65.0 & 79.3 & 0.0418 \\
        Non-Memo, Recite[1k], Guess, Others & 4 & 64.1 & 76.6 & 62.5 & 76.7 & 64.0 & 76.6 & 0.0418 \\
        Non-Memo, Recite[1k], Code, Others & 4 & 63.0 & 71.2 & 63.0 & 71.4 & 63.0 & 71.1 & 0.0503 \\
        Non-Memo, Recite[50], Reconstruct, Others & 4 & 62.7 & 79.2 & 62.5 & 79.4 & 63.0 & 79.3 & 0.0410 \\
        Non-Memo, Reconstruct, Recite[50], Others & 4 & 62.7 & 79.4 & 61.8 & 79.5 & 63.6 & 79.4 & 0.0394 \\
        Non-Memo, Recollect[1k], Reconstruct, Others & 4 & 62.7 & \underline{79.8} & 61.9 & 79.9 & 63.5 & \underline{79.8} & \textbf{0.0348} \\
        Non-Memo, Code, Recite[5], Others & 4 & 62.3 & 72.0 & 59.4 & 73.3 & 55.0 & 72.0 & 0.0501 \\
        Non-Memo, Guess, Recite[50], Others & 4 & 61.6 & 76.4 & 59.9 & 77.1 & 63.6 & 76.1 & 0.0437 \\
        Non-Memo, Guess, Recite[1k], Others & 4 & 60.1 & 76.5 & 63.1 & 76.3 & 57.3 & 76.8 & 0.0401 \\
        Non-Memo, Recite[1k], Reconstruct, Others & 4 & 59.8 & 77.7 & 61.2 & 77.7 & 58.6 & 78.2 & 0.0392 \\
        Non-Memo, Recollect[1k], Guess, Others & 4 & 58.4 & 76.9 & 64.9 & 76.9 & 53.2 & 77.4 & 0.0401 \\
        Non-Memo, Recollect[1k], Code, Others & 4 & 58.3 & 71.6 & 61.3 & 71.9 & 52.8 & 71.7 & 0.0489 \\
        Non-Memo, Code, Recite[50], Others & 4 & 57.7 & 68.1 & 51.4 & 69.4 & 52.6 & 67.8 & 0.0520 \\
        Non-Memo, Reconstruct, Recite[1k], Others & 4 & 57.6 & 77.6 & 60.0 & 77.4 & 55.5 & 77.9 & \underline{0.0386} \\
        Non-Memo, Code, Recite[1k], Others & 4 & 53.5 & 67.6 & 58.6 & 68.9 & 45.6 & 68.1 & 0.0513 \\
        Non-Memo, Recite[5], Code, Others & 4 & 49.1 & 69.1 & 52.8 & 69.4 & 45.9 & 69.3 & 0.0509 \\
        Non-Memo, Recite[50], Code, Others & 4 & 48.9 & 69.6 & 56.0 & 70.1 & 43.5 & 69.9 & 0.0519 \\
        \midrule
        $\bigstar$ \textbf{Non-Memo, Guess, Others} & 3 & \textbf{88.9} & \underline{89.9} & \underline{88.8} & \underline{89.9} & \textbf{88.6} & \underline{89.9} & \underline{0.0232} \\
        Non-Memo, Reconstruct, Others & 3 & \underline{87.6} & \textbf{91.0} & \textbf{89.3} & \textbf{91.1} & 82.9 & \textbf{91.1} & \textbf{0.0216} \\
        Non-Memo, Guess-or-Recollect[5], Others & 3 & 84.6 & 87.7 & 86.2 & 87.7 & 82.3 & 87.7 & 0.0275 \\
        Non-Memo, Reconstruct-or-Recollect[5], Others & 3 & 81.3 & 85.3 & 78.9 & 85.5 & \underline{83.7} & 85.2 & 0.0311 \\
        Non-Memo, Recite[5], Others & 3 & 80.4 & 84.5 & 77.2 & 84.8 & 82.5 & 84.4 & 0.0322 \\
        Non-Memo, Guess-or-Recollect[50], Others & 3 & 80.4 & 85.7 & 80.9 & 86.1 & 76.1 & 85.8 & 0.0290 \\
        Non-Memo, Recollect[5], Others & 3 & 79.5 & 83.6 & 77.0 & 83.7 & 80.4 & 83.5 & 0.0324 \\
        Non-Memo, Reconstruct-or-Recollect[50], Others & 3 & 78.1 & 84.0 & 78.5 & 84.0 & 77.7 & 84.0 & 0.0330 \\
        Non-Memo, Recite[5]-or-Reconstruct, Others & 3 & 77.6 & 82.3 & 72.1 & 83.0 & 76.0 & 82.2 & 0.0336 \\
        Non-Memo, Code, Others & 3 & 75.9 & 81.7 & 72.3 & 82.5 & 69.6 & 81.8 & 0.0331 \\
        Non-Memo, Recollect[50], Others & 3 & 75.4 & 82.3 & 76.2 & 82.5 & 74.7 & 82.3 & 0.0343 \\
        Non-Memo, Code-or-Recollect[5], Others & 3 & 75.2 & 81.3 & 73.1 & 81.5 & 76.2 & 81.3 & 0.0331 \\
        Non-Memo, Recite[50], Others & 3 & 75.2 & 82.3 & 77.8 & 82.4 & 72.7 & 82.4 & 0.0332 \\
        Non-Memo, Code-or-Recollect[50], Others & 3 & 74.1 & 80.5 & 71.5 & 80.6 & 73.7 & 80.4 & 0.0336 \\
        Non-Memo, Recite[5]-or-Code, Others & 3 & 72.7 & 78.3 & 66.2 & 79.5 & 67.1 & 78.2 & 0.0374 \\
        Non-Memo, Recite[50]-or-Reconstruct, Others & 3 & 72.5 & 79.7 & 69.2 & 80.0 & 74.6 & 79.5 & 0.0353 \\
        Non-Memo, Recite[1k]-or-Code, Others & 3 & 71.3 & 78.1 & 71.9 & 78.1 & 70.7 & 78.1 & 0.0363 \\
        Non-Memo, Recite[5]-or-Guess, Others & 3 & 70.6 & 77.0 & 63.3 & 78.5 & 66.9 & 76.7 & 0.0384 \\
        Non-Memo, Guess-or-Recollect[1k], Others & 3 & 70.6 & 80.7 & 74.4 & 80.9 & 67.2 & 80.8 & 0.0331 \\
        Non-Memo, Recollect[1k], Others & 3 & 68.5 & 79.1 & 70.1 & 79.3 & 67.1 & 79.0 & 0.0346 \\
        Non-Memo, Recite[50]-or-Code, Others & 3 & 68.0 & 75.3 & 60.7 & 77.4 & 59.2 & 75.1 & 0.0394 \\
        Non-Memo, Recite[1k], Others & 3 & 67.6 & 78.3 & 69.6 & 78.4 & 65.7 & 78.4 & 0.0344 \\
        Non-Memo, Recite[50]-or-Guess, Others & 3 & 67.2 & 75.1 & 61.3 & 76.3 & 66.6 & 74.7 & 0.0401 \\
        Non-Memo, Recite[1k]-or-Guess, Others & 3 & 66.8 & 76.0 & 63.9 & 77.5 & 58.3 & 76.2 & 0.0373 \\
        Non-Memo, Code-or-Recollect[1k], Others & 3 & 66.1 & 76.6 & 64.9 & 76.8 & 67.4 & 76.4 & 0.0368 \\
        Non-Memo, Reconstruct-or-Recollect[1k], Others & 3 & 65.5 & 78.0 & 70.7 & 78.7 & 60.5 & 78.2 & 0.0344 \\
        Non-Memo, Recite[1k]-or-Reconstruct, Others & 3 & 65.3 & 75.5 & 62.1 & 75.9 & 68.6 & 75.2 & 0.0369 \\
        \bottomrule
    \end{tabular}
    \caption{Same experiments as in Table~\ref{tab:taxonomy_results_full}, restricted to the Pythia 6.9B model. Our data-driven taxonomy, marked with $\bigstar$, also achieves the highest performance when this model is evaluated in isolation. Similar results for Pythia 12B and Pythia 2.8B are provided in Tables \ref{tab:taxonomy_results_12b} and \ref{tab:taxonomy_results_2-8b}.}
    \label{tab:taxonomy_results_6-9b}
\end{table*}

%% file: figures/tax_perf/tax_perf_2-8b.tex
\begin{table*}[!ht]
    \centering
    \small
    \begin{tabular}{@{}l@{}cccccccc@{}} 
        \toprule
        \textbf{Taxonomy name} & \textbf{Classes} & \textbf{Min $F_1$} & \textbf{Mean $F_1$} & \textbf{Min Prec} & \textbf{Mean Prec} & \textbf{Min Rec} & \textbf{Mean Rec} & \textbf{Mean Loss} \\
        \midrule
        Non-Memo, Recollect[50], Code, Others & 4 & \textbf{75.7} & 79.3 & \textbf{72.8} & 79.8 & 70.2 & 79.3 & 0.0433 \\
        Non-Memo, Recollect[5], Guess, Others & 4 & \underline{73.1} & 80.2 & 65.8 & 81.4 & 67.7 & 80.0 & 0.0403 \\
        Non-Memo, Recollect[5], Code, Others & 4 & 72.6 & 78.9 & \underline{70.6} & 80.1 & 67.4 & 78.9 & 0.0434 \\
        Non-Memo, Recollect[50], Guess, Others & 4 & 72.3 & 80.2 & 66.8 & 80.8 & 72.2 & 80.0 & 0.0402 \\
        Non-Memo, Recollect[50], Reconstruct, Others & 4 & 71.5 & \textbf{83.6} & 68.3 & \textbf{84.1} & 75.1 & \textbf{83.4} & \textbf{0.0358} \\
        Non-Memo, Recite[5], Guess, Others & 4 & 70.7 & 79.3 & 62.2 & 81.2 & 69.3 & 78.9 & 0.0427 \\
        Non-Memo, Guess, Recite[5], Others & 4 & 70.3 & 79.7 & 63.2 & 80.9 & \textbf{77.2} & 79.3 & 0.0423 \\
        Non-Memo, Recollect[5], Reconstruct, Others & 4 & 69.9 & \underline{81.8} & 63.7 & \underline{82.8} & \underline{75.9} & 81.4 & 0.0389 \\
        Non-Memo, Guess, Recite[50], Others & 4 & 68.6 & 79.7 & 63.4 & 80.5 & 74.6 & 79.4 & 0.0420 \\
        Non-Memo, Reconstruct, Recite[5], Others & 4 & 66.8 & 80.0 & 61.1 & 81.0 & 73.8 & 79.6 & 0.0410 \\
        Non-Memo, Recite[50], Guess, Others & 4 & 66.6 & 78.7 & 62.0 & 79.7 & 72.1 & 78.4 & 0.0431 \\
        Non-Memo, Recollect[1k], Reconstruct, Others & 4 & 66.6 & 81.7 & 64.1 & 82.0 & 69.3 & \underline{81.5} & \underline{0.0362} \\
        $\blacklozenge$ \textbf{Non-Memo, Recite[5], Reconstruct, Others} & 4 & 66.2 & 80.4 & 63.9 & 80.9 & 68.6 & 80.2 & 0.0399 \\
        Non-Memo, Code, Recite[5], Others & 4 & 64.7 & 73.4 & 60.0 & 75.1 & 61.5 & 73.2 & 0.0485 \\
        Non-Memo, Recite[1k], Guess, Others & 4 & 64.4 & 76.2 & 59.6 & 76.9 & 62.4 & 75.9 & 0.0423 \\
        Non-Memo, Reconstruct, Recite[50], Others & 4 & 63.3 & 79.8 & 63.5 & 80.0 & 63.0 & 79.8 & 0.0408 \\
        Non-Memo, Recite[50], Reconstruct, Others & 4 & 63.1 & 79.8 & 63.4 & 80.2 & 62.8 & 79.8 & 0.0410 \\
        Non-Memo, Recollect[1k], Guess, Others & 4 & 62.5 & 77.7 & 63.7 & 78.0 & 61.4 & 77.6 & 0.0410 \\
        Non-Memo, Recollect[1k], Code, Others & 4 & 62.2 & 73.5 & 63.3 & 73.9 & 61.1 & 73.4 & 0.0474 \\
        Non-Memo, Recite[1k], Code, Others & 4 & 61.5 & 70.3 & 58.9 & 71.2 & 56.1 & 70.1 & 0.0505 \\
        Non-Memo, Guess, Recite[1k], Others & 4 & 61.5 & 76.7 & 61.5 & 77.0 & 61.5 & 76.8 & 0.0409 \\
        Non-Memo, Code, Recite[50], Others & 4 & 60.3 & 70.5 & 56.0 & 72.2 & 53.8 & 70.3 & 0.0505 \\
        Non-Memo, Recite[1k], Reconstruct, Others & 4 & 59.7 & 78.7 & 64.8 & 78.6 & 55.3 & 79.2 & 0.0399 \\
        Non-Memo, Reconstruct, Recite[1k], Others & 4 & 59.6 & 78.1 & 60.6 & 78.2 & 58.5 & 78.2 & 0.0393 \\
        Non-Memo, Code, Recite[1k], Others & 4 & 53.1 & 67.8 & 58.2 & 69.2 & 45.8 & 68.4 & 0.0512 \\
        Non-Memo, Recite[5], Code, Others & 4 & 50.7 & 70.7 & 55.3 & 72.6 & 43.8 & 70.8 & 0.0500 \\
        Non-Memo, Recite[50], Code, Others & 4 & 46.7 & 69.5 & 54.1 & 70.2 & 41.0 & 70.0 & 0.0506 \\
        \midrule
        $\bigstar$ \textbf{Non-Memo, Guess, Others} & 3 & \textbf{88.4} & \underline{89.7} & \underline{87.0} & \underline{89.8} & \textbf{89.4} & \underline{89.7} & \underline{0.0249} \\
        Non-Memo, Reconstruct, Others & 3 & \underline{88.1} & \textbf{90.9} & \textbf{89.6} & \textbf{91.1} & 83.8 & \textbf{91.0} & \textbf{0.0230} \\
        Non-Memo, Recollect[5], Others & 3 & 83.2 & 86.9 & 78.5 & 87.4 & \underline{85.6} & 86.8 & 0.0294 \\
        Non-Memo, Reconstruct-or-Recollect[5], Others & 3 & 83.1 & 86.9 & 79.7 & 87.1 & 85.4 & 86.8 & 0.0297 \\
        Non-Memo, Guess-or-Recollect[5], Others & 3 & 83.0 & 87.0 & 82.3 & 87.0 & 83.7 & 86.9 & 0.0286 \\
        Non-Memo, Guess-or-Recollect[50], Others & 3 & 82.4 & 87.0 & 81.8 & 87.2 & 82.9 & 87.0 & 0.0286 \\
        Non-Memo, Recite[5], Others & 3 & 81.9 & 86.0 & 78.1 & 86.3 & 84.3 & 85.9 & 0.0292 \\
        Non-Memo, Recollect[50], Others & 3 & 80.6 & 85.8 & 78.2 & 86.0 & 83.2 & 85.7 & 0.0298 \\
        Non-Memo, Reconstruct-or-Recollect[50], Others & 3 & 80.2 & 85.6 & 78.0 & 86.0 & 82.7 & 85.5 & 0.0300 \\
        Non-Memo, Recite[50], Others & 3 & 79.8 & 85.3 & 78.7 & 85.6 & 81.0 & 85.3 & 0.0305 \\
        Non-Memo, Recite[5]-or-Reconstruct, Others & 3 & 78.9 & 83.7 & 72.7 & 84.5 & 79.2 & 83.4 & 0.0326 \\
        Non-Memo, Recite[50]-or-Reconstruct, Others & 3 & 76.4 & 82.4 & 71.0 & 83.2 & 77.8 & 82.2 & 0.0334 \\
        Non-Memo, Code-or-Recollect[5], Others & 3 & 76.3 & 81.8 & 70.7 & 82.5 & 72.8 & 81.6 & 0.0325 \\
        Non-Memo, Recite[5]-or-Code, Others & 3 & 75.0 & 80.2 & 66.7 & 82.0 & 70.0 & 79.9 & 0.0362 \\
        Non-Memo, Code, Others & 3 & 74.6 & 80.9 & 70.0 & 82.2 & 68.3 & 80.8 & 0.0349 \\
        Non-Memo, Code-or-Recollect[50], Others & 3 & 74.0 & 80.9 & 70.5 & 81.3 & 74.8 & 80.7 & 0.0333 \\
        Non-Memo, Recite[5]-or-Guess, Others & 3 & 72.7 & 78.0 & 62.5 & 81.1 & 65.3 & 77.6 & 0.0381 \\
        Non-Memo, Recite[1k], Others & 3 & 72.3 & 80.7 & 69.4 & 81.1 & 75.5 & 80.4 & 0.0333 \\
        Non-Memo, Recite[50]-or-Code, Others & 3 & 71.6 & 77.4 & 62.2 & 80.0 & 64.8 & 77.1 & 0.0379 \\
        Non-Memo, Recite[50]-or-Guess, Others & 3 & 71.2 & 77.3 & 62.8 & 79.4 & 65.8 & 76.9 & 0.0390 \\
        Non-Memo, Recite[1k]-or-Code, Others & 3 & 70.4 & 77.6 & 68.6 & 77.7 & 71.3 & 77.5 & 0.0373 \\
        Non-Memo, Recite[1k]-or-Guess, Others & 3 & 70.0 & 76.8 & 67.1 & 77.0 & 66.9 & 76.8 & 0.0367 \\
        Non-Memo, Reconstruct-or-Recollect[1k], Others & 3 & 69.9 & 79.9 & 71.6 & 79.8 & 68.3 & 79.9 & 0.0335 \\
        Non-Memo, Code-or-Recollect[1k], Others & 3 & 69.5 & 78.3 & 65.9 & 78.8 & 70.6 & 78.1 & 0.0355 \\
        Non-Memo, Recollect[1k], Others & 3 & 69.4 & 79.7 & 71.7 & 80.1 & 67.1 & 79.7 & 0.0338 \\
        Non-Memo, Guess-or-Recollect[1k], Others & 3 & 68.7 & 80.1 & 73.2 & 80.6 & 63.0 & 80.4 & 0.0326 \\
        Non-Memo, Recite[1k]-or-Reconstruct, Others & 3 & 67.7 & 76.4 & 62.7 & 77.2 & 65.2 & 76.2 & 0.0365 \\
        \bottomrule
    \end{tabular}
    \caption{Same experiments as in Table~\ref{tab:taxonomy_results_full}, restricted to the Pythia 2.8B model. Our data-driven taxonomy, marked with $\bigstar$, also achieves the highest performance when this model is evaluated in isolation. Similar results for Pythia 12B and Pythia 6.9B are provided in Tables \ref{tab:taxonomy_results_12b} and \ref{tab:taxonomy_results_6-9b}.}
    \label{tab:taxonomy_results_2-8b}
\end{table*}

%% file: figures/confusion_matrices/full_confusion_4_classes.tex
\begin{figure*}[t!]
    \centering
    \includegraphics[width=\textwidth]{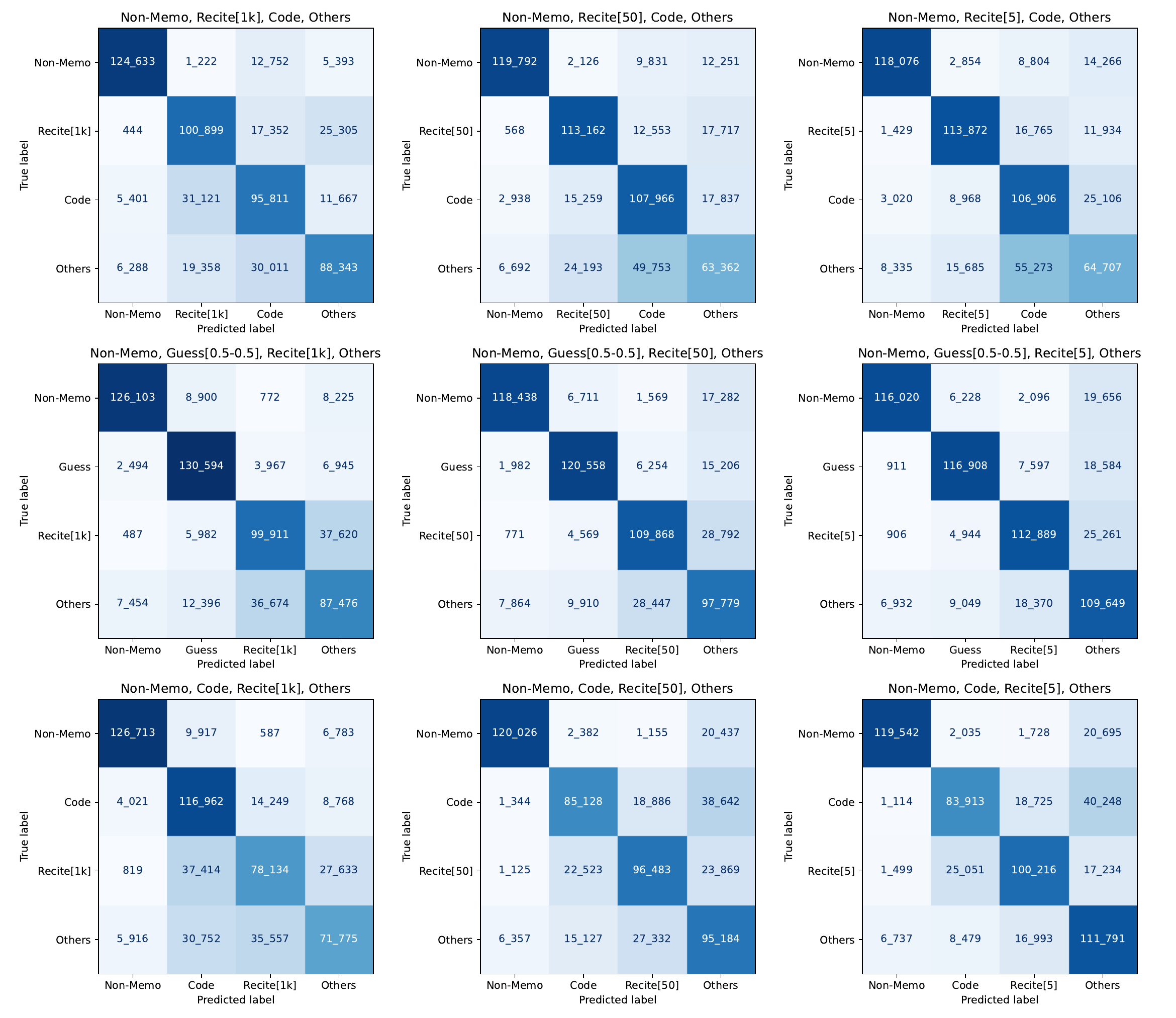}
    \caption{Confusion matrices of CNN classifiers trained with 4-classes taxonomies (part 1/3)}
    \label{fig:full_confusion_4_classes_part_1}
\end{figure*}

\begin{figure*}[t!]
    \centering
    \includegraphics[width=\textwidth]{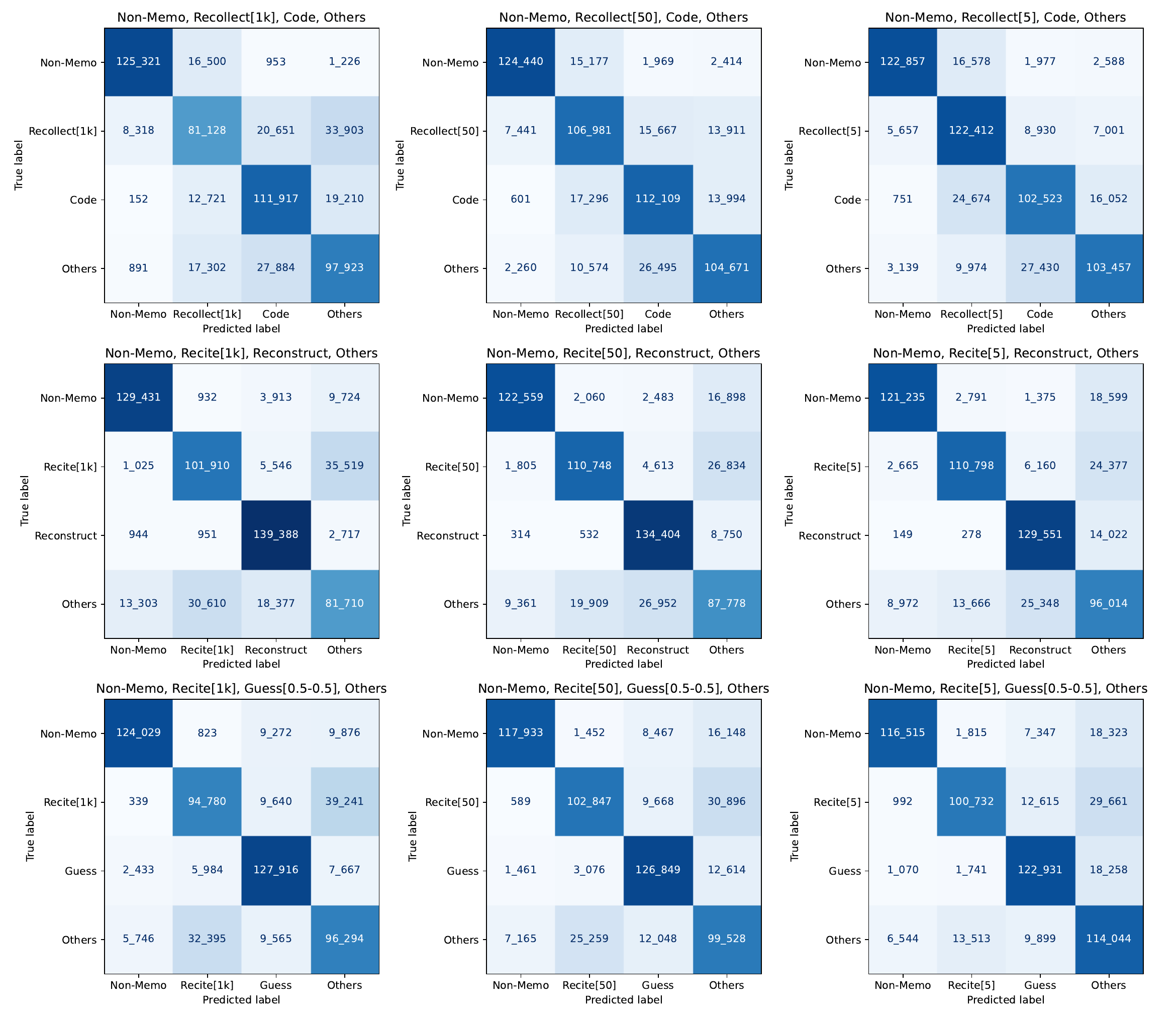}
    \caption{Confusion matrices of CNN classifiers trained with 4-classes taxonomies (part 2/3).}
    \label{fig:full_confusion_4_classes_part_2}
\end{figure*}

\begin{figure*}[t!]
    \centering
    \includegraphics[width=\textwidth]{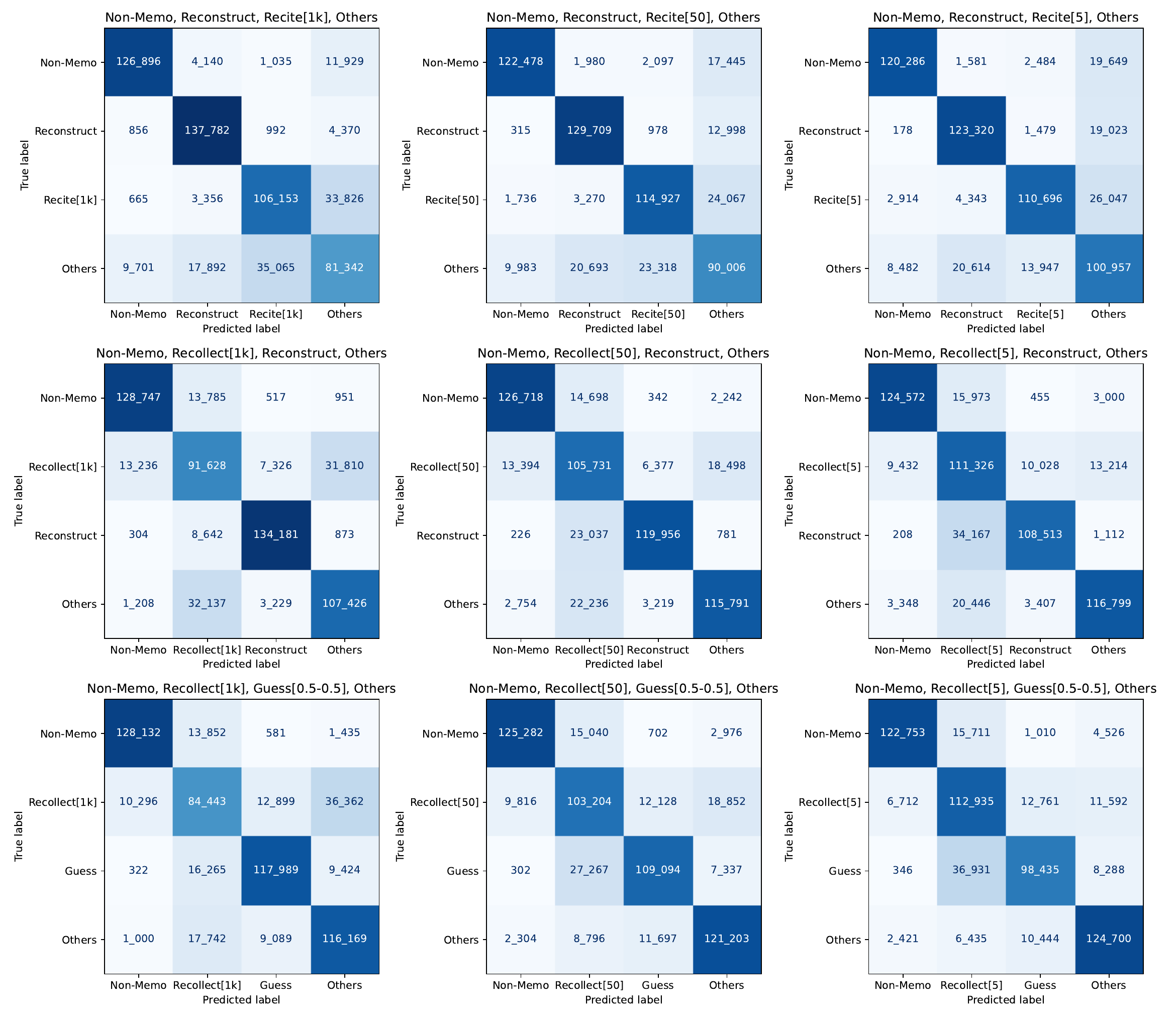}
    \caption{Confusion matrices of CNN classifiers trained with 4-classes taxonomies (part 3/3).}
    \label{fig:full_confusion_4_classes_part_3}
\end{figure*}

%% file: figures/confusion_matrices/full_confusion_3_classes.tex
\begin{figure*}[t!]
    \centering
    \includegraphics[width=\textwidth]{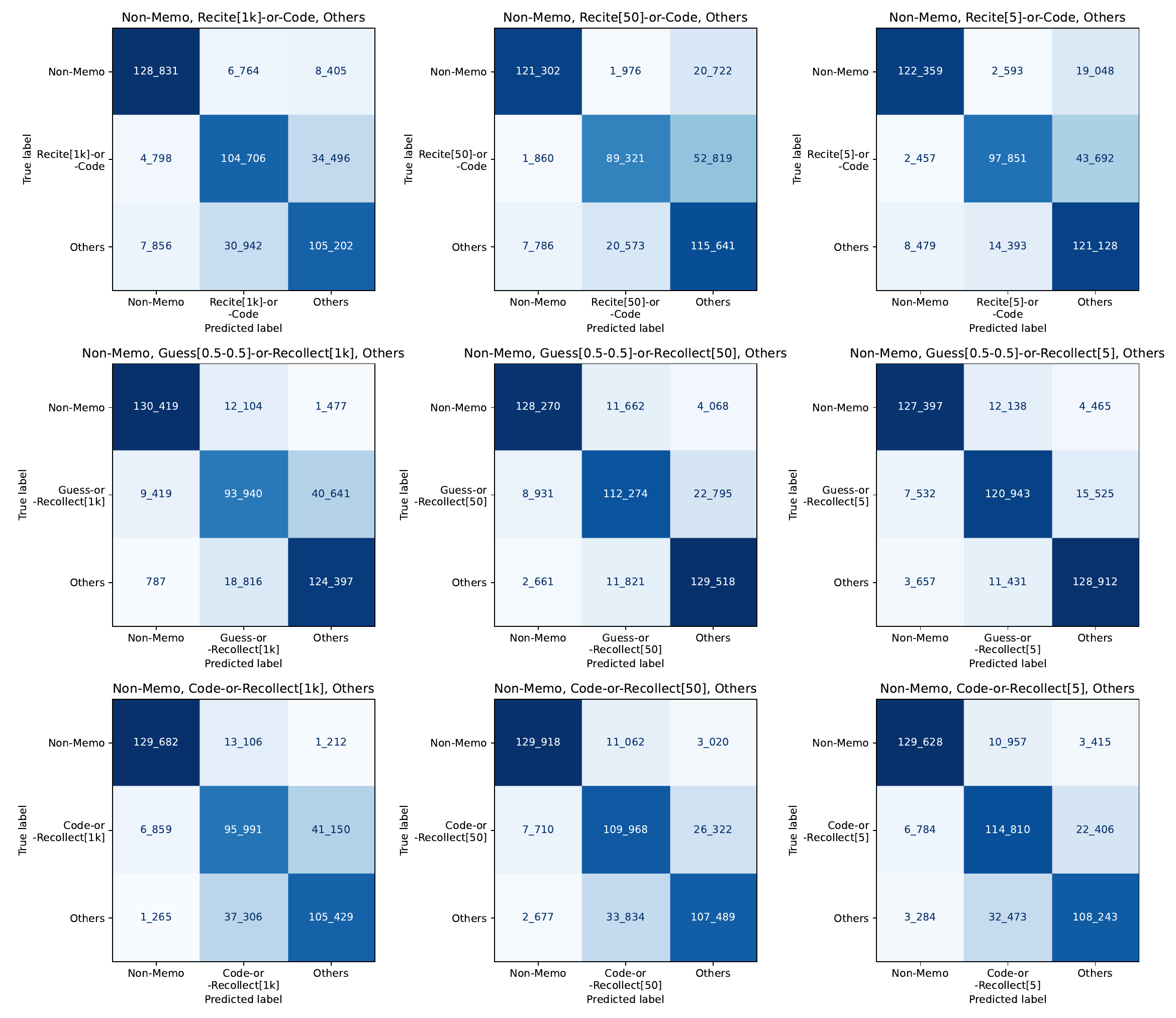}
    \caption{Confusion matrices of CNN classifiers trained with 3-classes taxonomies (part 1/3).}
    \label{fig:full_confusion_3_classes_part_1}
\end{figure*}

\begin{figure*}[t!]
    \centering
    \includegraphics[width=\textwidth]{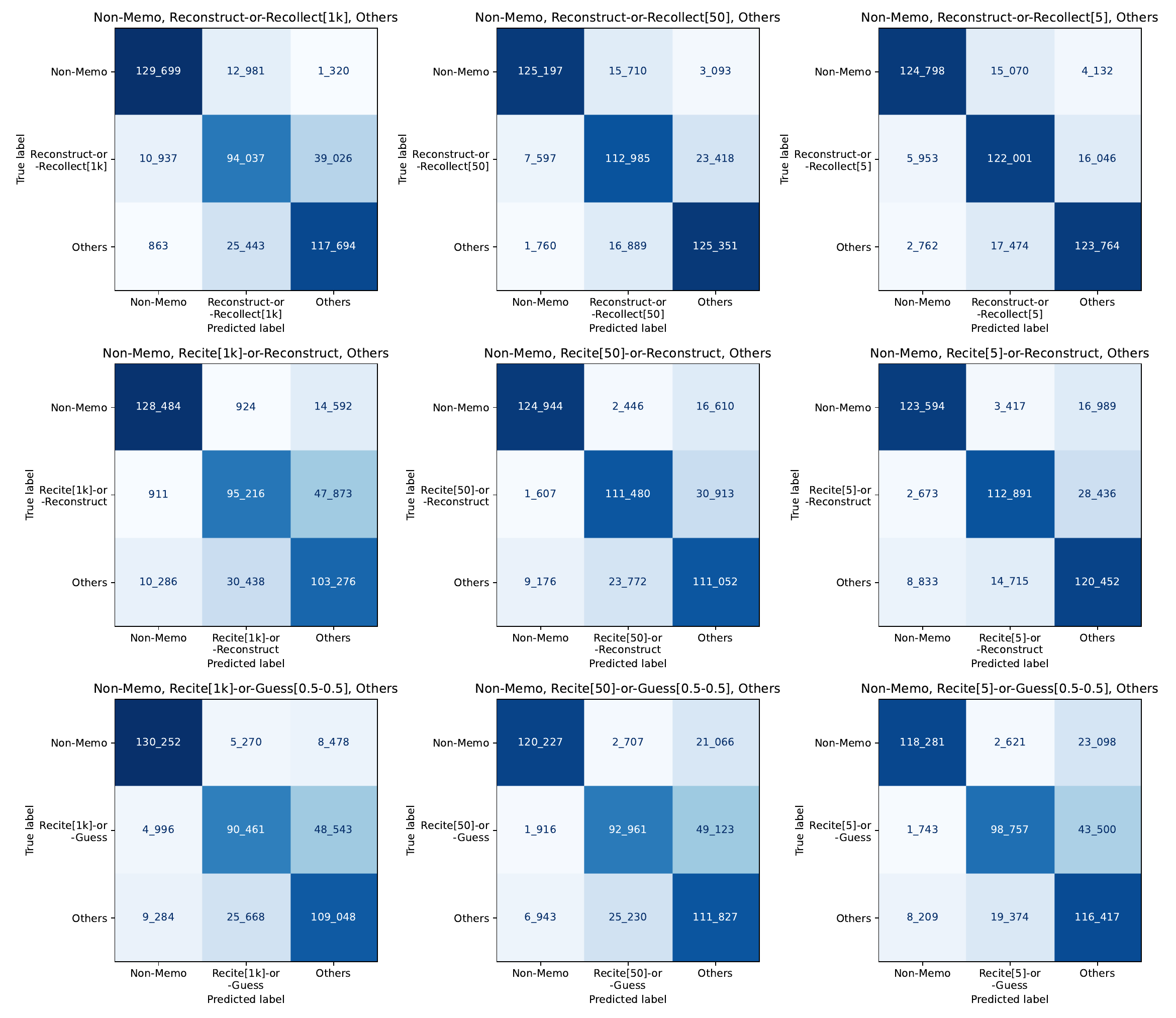}
    \caption{Confusion matrices of CNN classifiers trained with 3-classes taxonomies (part 2/3).}
    \label{fig:full_confusion_3_classes_part_2}
\end{figure*}

\begin{figure*}[t!]
    \centering
    \includegraphics[width=\textwidth]{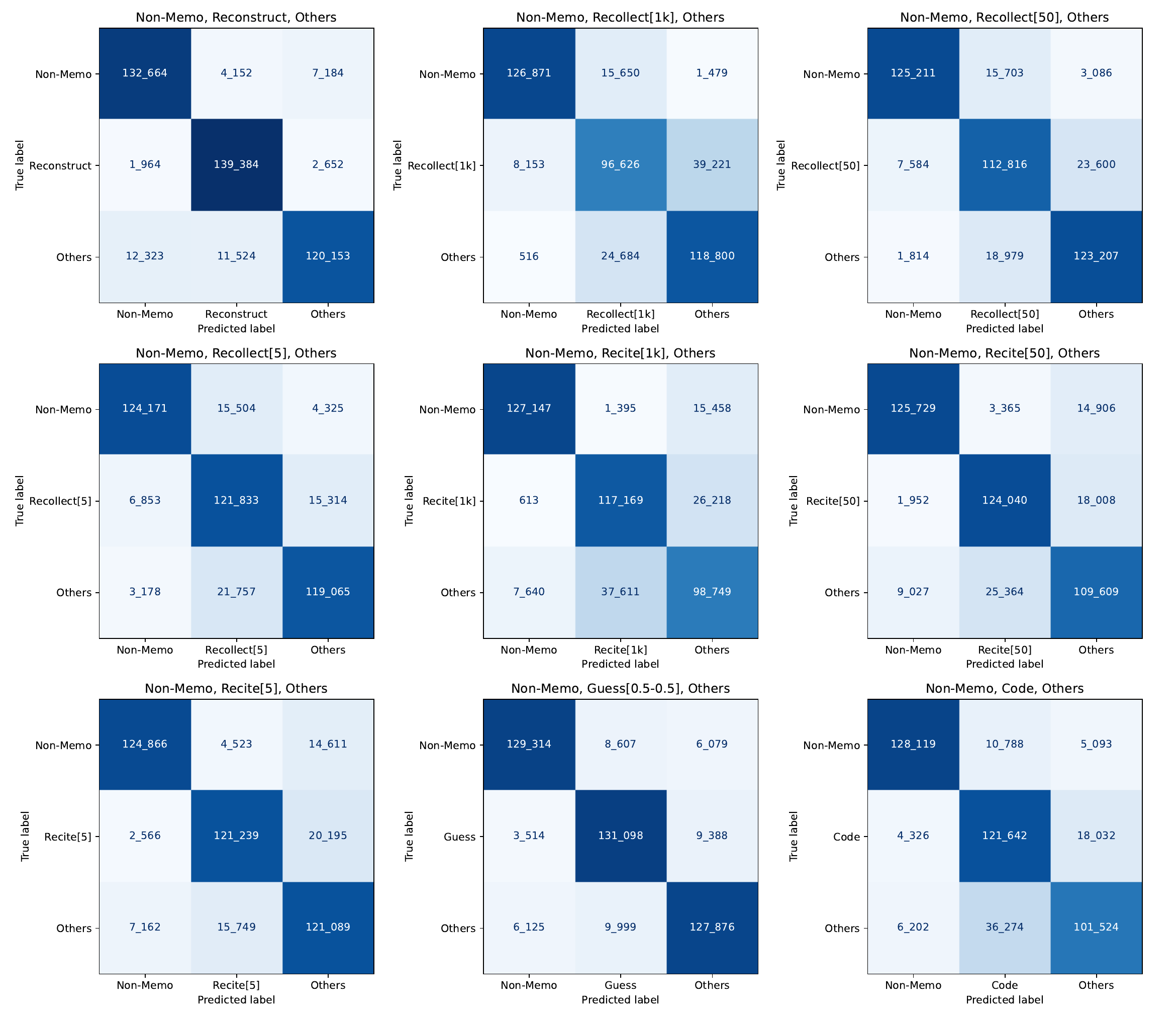}
    \caption{Confusion matrices of CNN classifiers trained with 3-classes taxonomies (part 3/3).}
    \label{fig:full_confusion_3_classes_part_3}
\end{figure*}

%% file: figures/delta_matrices/full_delta.tex
\begin{figure*}[t!]
    \centering
    \includegraphics[height=0.95\textheight]{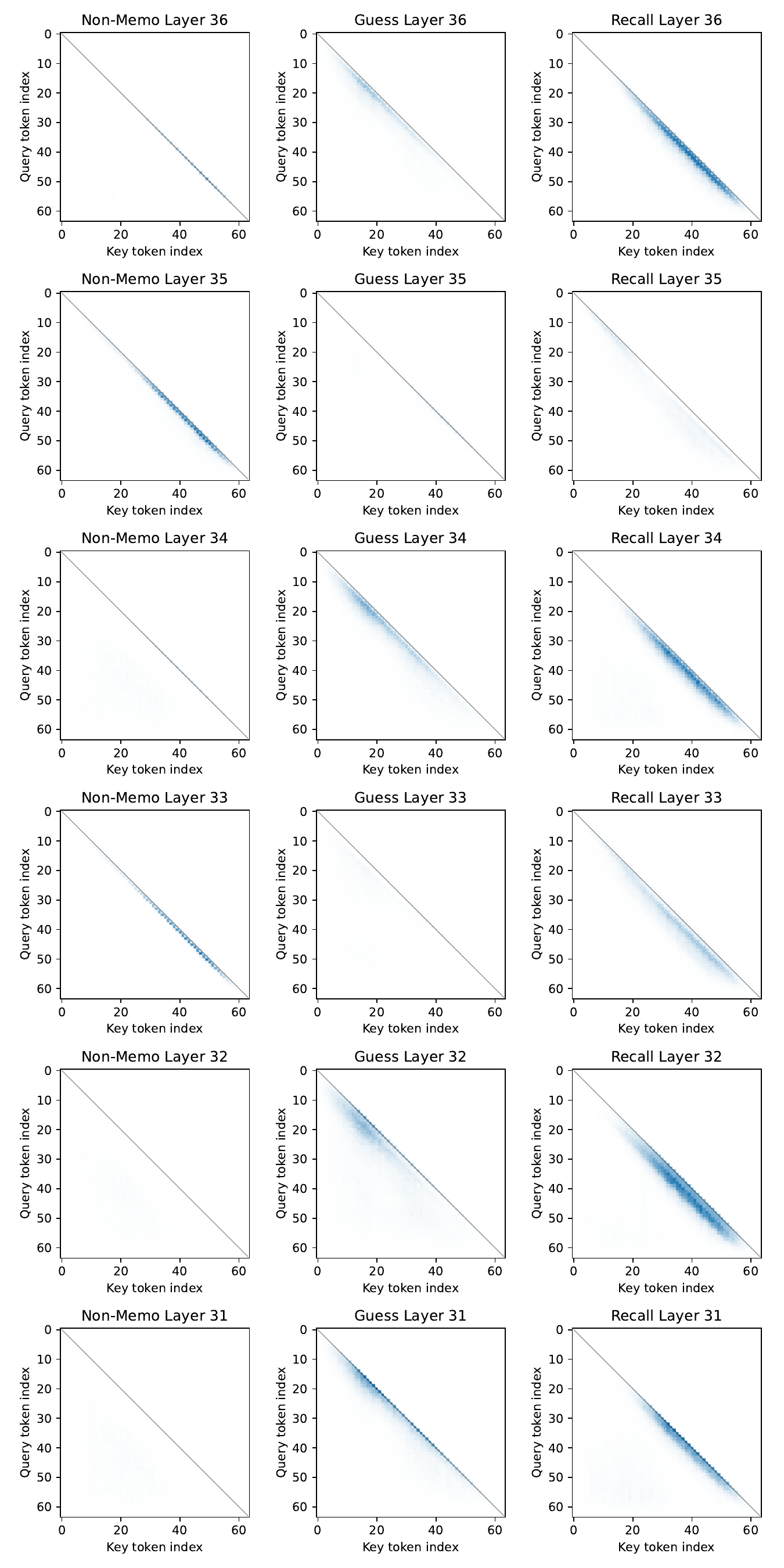}
    \cprotect\caption{Visualizing $\Delta_l[\verb|Non-Memo|]$ and  $\Delta_l[\verb|Guess|]$ and $\Delta_l[\verb|Recall|]$ for $l \in \llbracket 31; 36 \rrbracket$.}
    \label{fig:full_delta_matrix_31_36}
\end{figure*}

\begin{figure*}[t!]
    \centering
    \includegraphics[height=0.95\textheight]{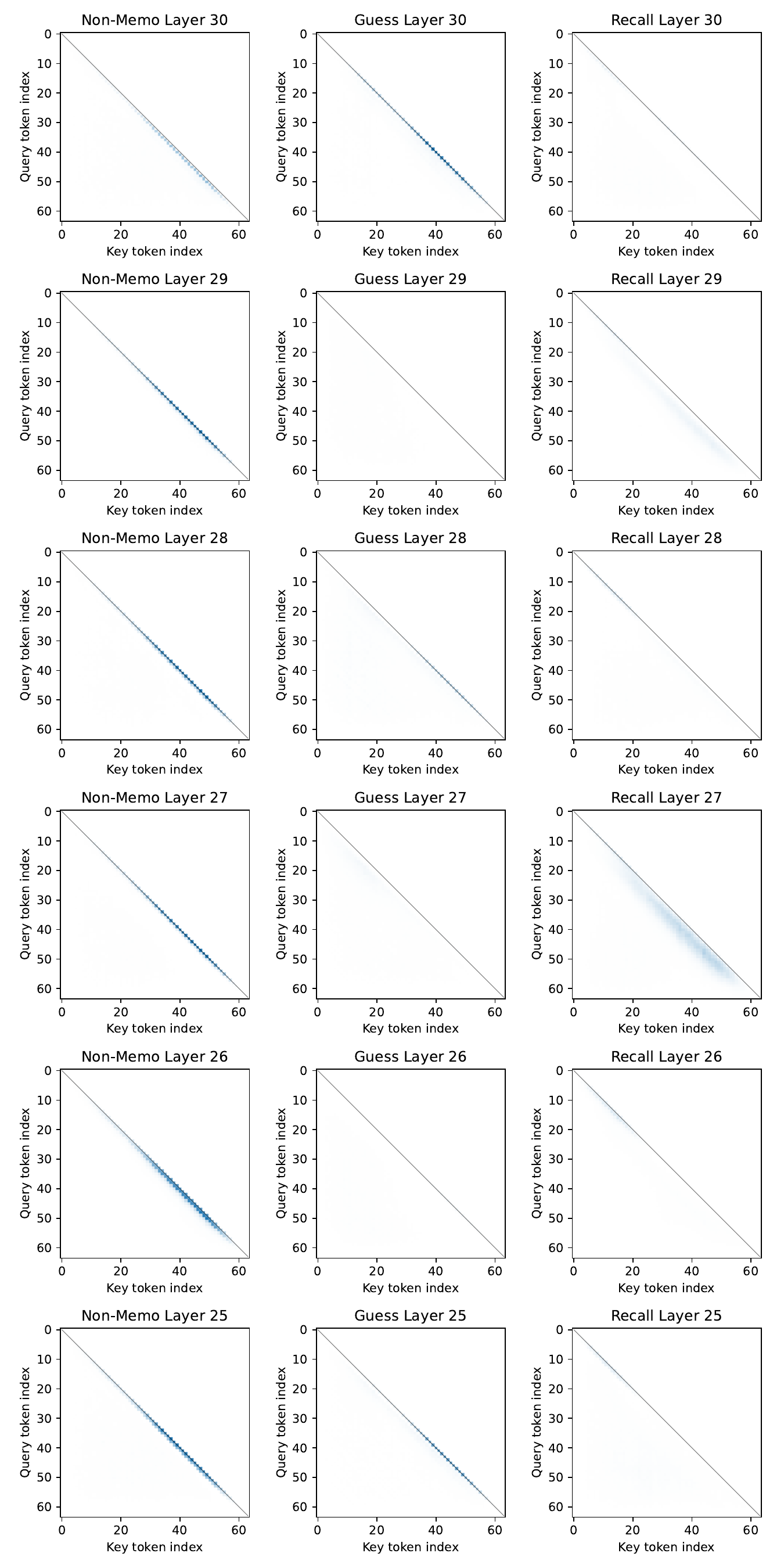}
    \cprotect\caption{Visualizing $\Delta_l[\verb|Non-Memo|]$ and  $\Delta_l[\verb|Guess|]$ and $\Delta_l[\verb|Recall|]$ for $l \in \llbracket 25; 30 \rrbracket$.}
    \label{fig:full_delta_matrix_25_30}
\end{figure*}

\begin{figure*}[t!]
    \centering
    \includegraphics[height=0.95\textheight]{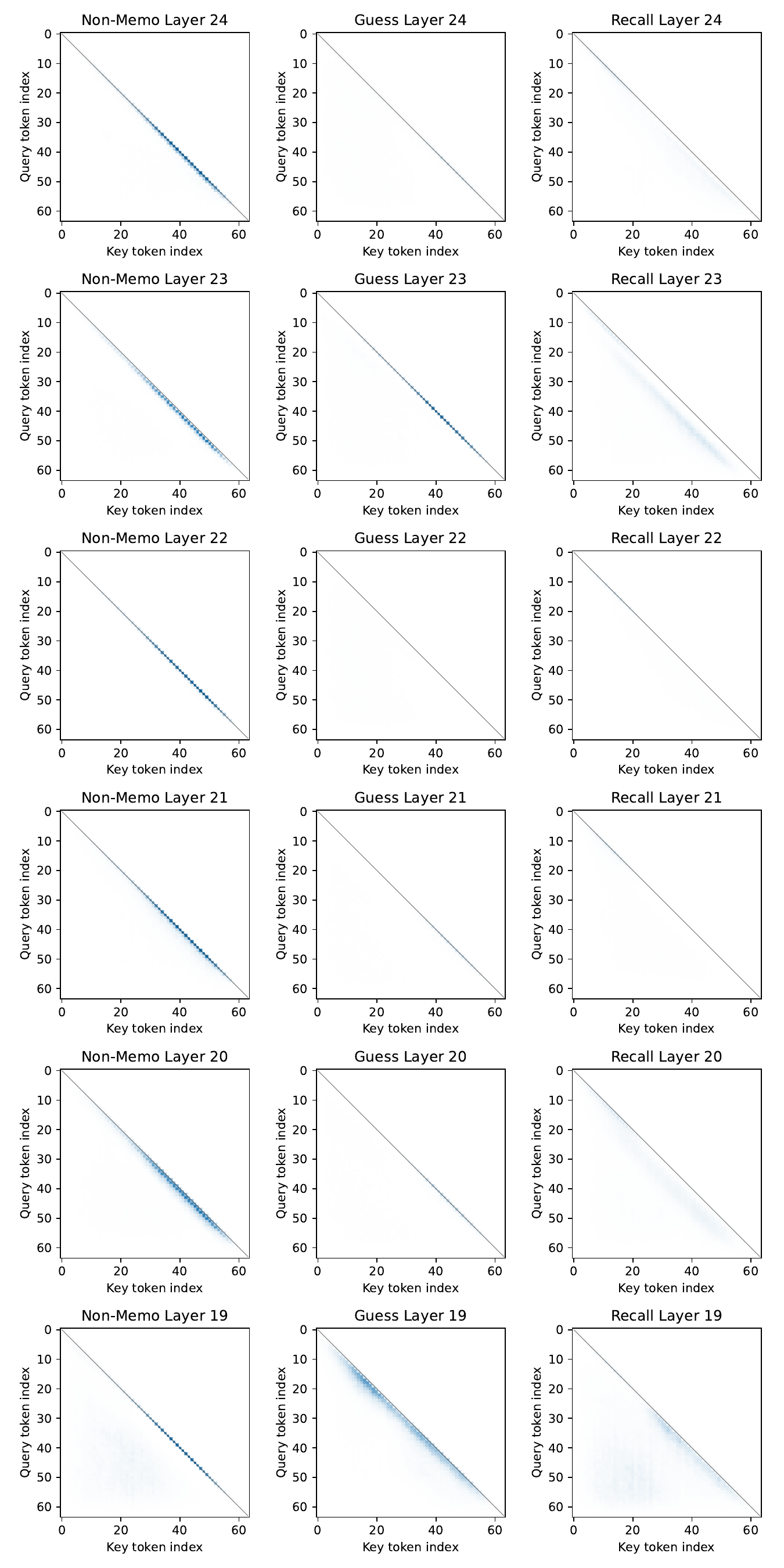}
    \cprotect\caption{Visualizing $\Delta_l[\verb|Non-Memo|]$ and  $\Delta_l[\verb|Guess|]$ and $\Delta_l[\verb|Recall|]$ for $l \in \llbracket 19; 24 \rrbracket$.}
    \label{fig:full_delta_matrix_19_24}
\end{figure*}

\begin{figure*}[t!]
    \centering
    \includegraphics[height=0.95\textheight]{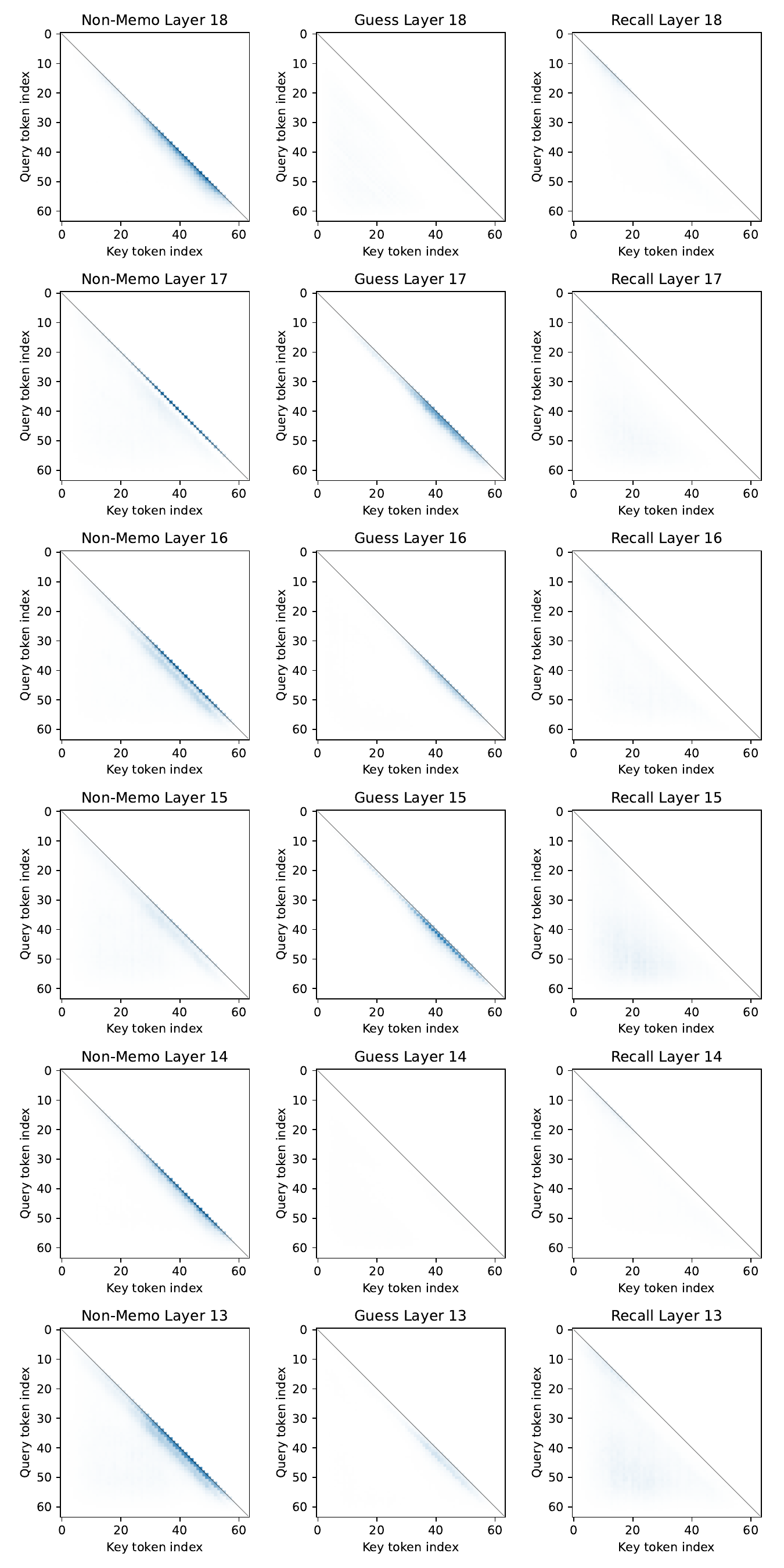}
    \cprotect\caption{Visualizing $\Delta_l[\verb|Non-Memo|]$ and  $\Delta_l[\verb|Guess|]$ and $\Delta_l[\verb|Recall|]$ for $l \in \llbracket 13; 18 \rrbracket$.}
    \label{fig:full_delta_matrix_13_18}
\end{figure*}

\begin{figure*}[t!]
    \centering
    \includegraphics[height=0.95\textheight]{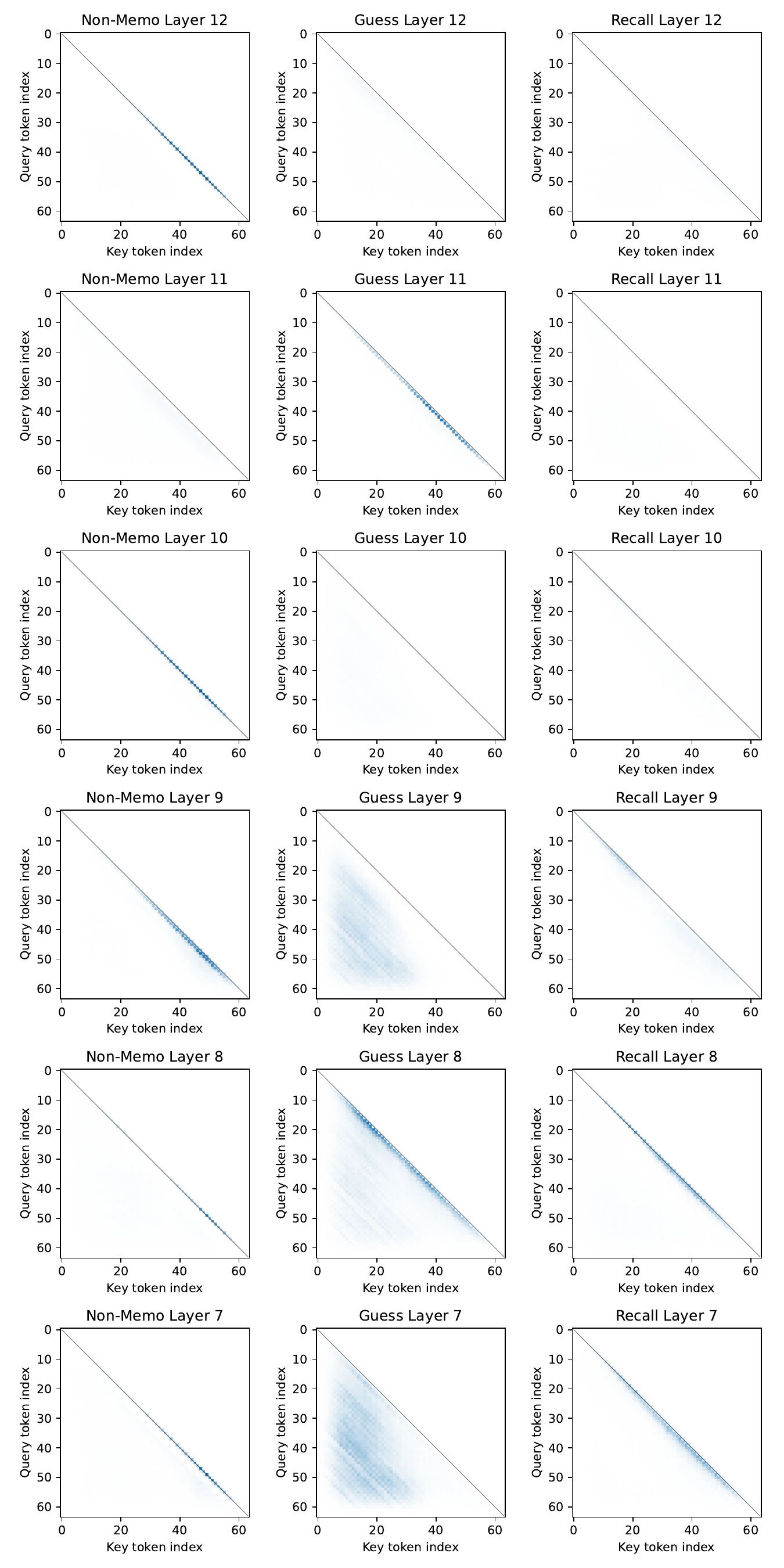}
    \cprotect\caption{Visualizing $\Delta_l[\verb|Non-Memo|]$ and  $\Delta_l[\verb|Guess|]$ and $\Delta_l[\verb|Recall|]$ for $l \in \llbracket 7; 12 \rrbracket$.}
    \label{fig:full_delta_matrix_7_12}
\end{figure*}

\begin{figure*}[t!]
    \centering
    \includegraphics[height=0.95\textheight]{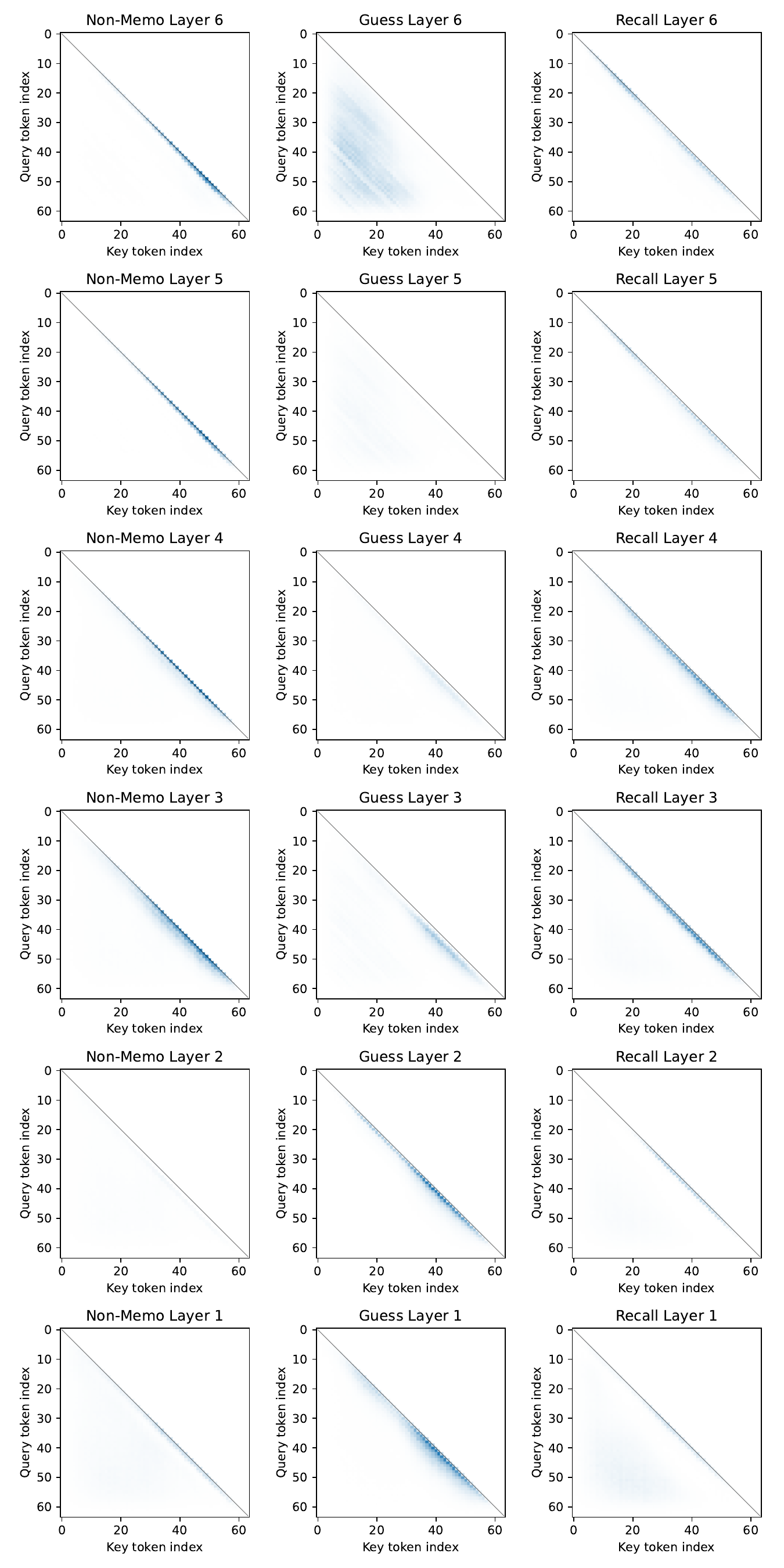}
    \cprotect\caption{Visualizing $\Delta_l[\verb|Non-Memo|]$ and  $\Delta_l[\verb|Guess|]$ and $\Delta_l[\verb|Recall|]$ for $l \in \llbracket 1; 6 \rrbracket$.}
    \label{fig:full_delta_matrix_1_6}
\end{figure*}

%% file: bib/custom.bib
@inproceedings{mireshghallah_empirical_2022,
	title = {An {Empirical} {Analysis} of {Memorization} in {Fine}-tuned {Autoregressive} {Language} {Models}},
	doi = {10.18653/v1/2022.emnlp-main.119},
	booktitle = {{ACL}-{EMNLP}},
	author = {Mireshghallah, Fatemehsadat and Uniyal, Archit and Wang, Tianhao and Evans, David and Berg-Kirkpatrick, Taylor},
	month = dec,
	year = {2022},
	pages = {1816--1826},
}

@inproceedings{shokri_membership_2017,
	title = {Membership {Inference} {Attacks} against {Machine} {Learning} {Models}},
	doi = {10.1109/SP.2017.41},
	booktitle = {{IEEE} {S}\&{P}},
	author = {Shokri, Reza and Stronati, Marco and Song, Congzheng and Shmatikov, Vitaly},
	year = {2017},
}

@inproceedings{yu_bag_2023,
	title = {Bag of {Tricks} for {Training} {Data} {Extraction} from {Language} {Models}},
	booktitle = {{ICML}},
	author = {Yu, Weichen and Pang, Tianyu and Liu, Qian and Du, Chao and Kang, Bingyi and Huang, Yan and Lin, Min and Yan, Shuicheng},
	month = jun,
	year = {2023},
}

@inproceedings{lee_language_2023,
	title = {Do {Language} {Models} {Plagiarize}?},
	doi = {10.1145/3543507.3583199},
	booktitle = {{ACM} {WWW}},
	author = {Lee, Jooyoung and Le, Thai and Chen, Jinghui and Lee, Dongwon},
	year = {2023},
	pages = {3637--3647},
}

@inproceedings{fredrikson_privacy_2014,
	title = {Privacy in {Pharmacogenetics}: {An} {End}-to-{End} {Case} {Study} of {Personalized} {Warfarin} {Dosing}},
	booktitle = {{USENIX} {Security}},
	author = {Fredrikson, Matthew and Lantz, Eric and Jha, Somesh and Lin, Simon and Page, David and Ristenpart, Thomas},
	year = {2014},
}

@inproceedings{feldman_does_2020,
	title = {Does learning require memorization? a short tale about a long tail},
	doi = {10.1145/3357713.3384290},
	booktitle = {{ACM} {SIGACT} {STOC}},
	author = {Feldman, Vitaly},
	month = jun,
	year = {2020},
	pages = {954--959},
}

@inproceedings{carlini_extracting_2023,
	title = {Extracting {Training} {Data} from {Diffusion} {Models}},
	booktitle = {{USENIX} {Security}},
	author = {Carlini, Nicholas and Hayes, Jamie and Nasr, Milad and Jagielski, Matthew and Sehwag, Vikash and Tramèr, Florian and Balle, Borja and Ippolito, Daphne and Wallace, Eric},
	month = aug,
	year = {2023},
}

@misc{black_gpt-neo_2021,
	title = {{GPT}-{Neo}: {Large} {Scale} {Autoregressive} {Language} {Modeling} with {Mesh}-{Tensorflow}},
	publisher = {Zenodo},
	author = {Black, Sid and Leo, Gao and Wang, Phil and Leahy, Connor and Biderman, Stella},
	month = mar,
	year = {2021},
}

@misc{mahloujifar_membership_2021,
	title = {Membership {Inference} on {Word} {Embedding} and {Beyond}},
	author = {Mahloujifar, Saeed and Inan, Huseyin A. and Chase, Melissa and Ghosh, Esha and Hasegawa, Marcello},
	month = jun,
	year = {2021},
	note = {arXiv:2106.11384},
	doi = {10.48550/arXiv.2106.11384},
}

@inproceedings{carlini_extracting_2021,
	title = {Extracting {Training} {Data} from {Large} {Language} {Models}.},
	booktitle = {{USENIX} {Security}},
	author = {Carlini, Nicholas and Tramer, Florian and Wallace, Eric and Jagielski, Matthew and Herbert-Voss, Ariel and Lee, Katherine and Roberts, Adam and Brown, Tom B and Song, Dawn and Erlingsson, Ulfar and Oprea, Alina and Raffel, Colin},
	year = {2021},
}

@inproceedings{carlini_quantifying_2023,
	title = {Quantifying {Memorization} {Across} {Neural} {Language} {Models}},
	booktitle = {{ICLR}},
	author = {Carlini, Nicholas and Ippolito, Daphne and Jagielski, Matthew and Lee, Katherine and Tramer, Florian and Zhang, Chiyuan},
	year = {2023},
}

@inproceedings{feldman_what_2020,
	title = {What {Neural} {Networks} {Memorize} and {Why}: {Discovering} the {Long} {Tail} via {Influence} {Estimation}},
	shorttitle = {What {Neural} {Networks} {Memorize} and {Why}},
	booktitle = {{NeurIPS}},
	author = {Feldman, Vitaly and Zhang, Chiyuan},
	year = {2020},
}

@inproceedings{nasr_scalable_2025,
	title = {Scalable {Extraction} of {Training} {Data} from ({Production}) {Language} {Models}},
	booktitle = {{ICLR}},
	author = {Nasr, Milad and Carlini, Nicholas and Hayase, Jonathan and Jagielski, Matthew and Cooper, A. Feder and Ippolito, Daphne and Choquette-Choo, Christopher A. and Wallace, Eric and Tramèr, Florian and Lee, Katherine},
	year = {2025},
}

@inproceedings{ippolito_preventing_2023,
	title = {Preventing {Generation} of {Verbatim} {Memorization} in {Language} {Models} {Gives} a {False} {Sense} of {Privacy}},
	doi = {10.18653/v1/2023.inlg-main.3},
	booktitle = {{INLG}},
	author = {Ippolito, Daphne and Tramer, Florian and Nasr, Milad and Zhang, Chiyuan and Jagielski, Matthew and Lee, Katherine and Choquette Choo, Christopher and Carlini, Nicholas},
	year = {2023},
	pages = {28--53},
}

@inproceedings{carlini_membership_2022,
	title = {Membership {Inference} {Attacks} {From} {First} {Principles}},
	doi = {10.1109/SP46214.2022.9833649},
	booktitle = {{IEEE} {S}\&{P}},
	author = {Carlini, Nicholas and Chien, Steve and Nasr, Milad and Song, Shuang and Terzis, Andreas and Tramèr, Florian},
	year = {2022},
	pages = {1897--1914},
}

@inproceedings{biderman_emergent_2023,
	title = {Emergent and {Predictable} {Memorization} in {Large} {Language} {Models}},
	volume = {36},
	booktitle = {{NeurIPS}},
	author = {Biderman, Stella and Prashanth, Usvsn Sai and Sutawika, Lintang and Schoelkopf, Hailey and Anthony, Quentin and Purohit, Shivanshu and Raff, Edward},
	year = {2023},
	pages = {28072--28090},
}

@misc{biderman_pythia_2023,
	title = {Pythia: {A} {Suite} for {Analyzing} {Large} {Language} {Models} {Across} {Training} and {Scaling}},
	shorttitle = {Pythia},
	publisher = {arXiv},
	author = {Biderman, Stella and Schoelkopf, Hailey and Anthony, Quentin and Bradley, Herbie and O'Brien, Kyle and Hallahan, Eric and Khan, Mohammad Aflah and Purohit, Shivanshu and Prashanth, Usvsn Sai and Raff, Edward and Skowron, Aviya and Sutawika, Lintang and van der Wal, Oskar},
	month = may,
	year = {2023},
	note = {arXiv:2304.01373},
	doi = {10.48550/arXiv.2304.01373},
}

@inproceedings{zhang_counterfactual_2023,
	title = {Counterfactual {Memorization} in {Neural} {Language} {Models}},
	booktitle = {{NeurIPS}},
	author = {Zhang, Chiyuan and Ippolito, Daphne and Lee, Katherine and Jagielski, Matthew and Tramèr, Florian and Carlini, Nicholas},
	year = {2023},
}

@inproceedings{zhang_understanding_2017,
	title = {Understanding deep learning requires rethinking generalization},
	booktitle = {{ICLR}},
	author = {Zhang, Chiyuan and Bengio, Samy and Hardt, Moritz and Recht, Benjamin and Vinyals, Oriol},
	year = {2017},
}

@inproceedings{meeus_copyright_2024,
	title = {Copyright {Traps} for {Large} {Language} {Models}},
	booktitle = {{ICML}},
	author = {Meeus, Matthieu and Shilov, Igor and Faysse, Manuel and de Montjoye, Yves-Alexandre},
	year = {2024},
}

@inproceedings{dentan_reconstructing_2024,
	title = {Reconstructing training data from document understanding models},
	booktitle = {{USENIX} {Security}},
	author = {Dentan, Jérémie and Paran, Arnaud and Shabou, Aymen},
	year = {2024},
	pages = {6813--6830},
}

@inproceedings{huang_demystifying_2024,
	title = {Demystifying {Verbatim} {Memorization} in {Large} {Language} {Models}},
	doi = {10.18653/v1/2024.emnlp-main.598},
	booktitle = {{EMNLP}},
	author = {Huang, Jing and Yang, Diyi and Potts, Christopher},
	year = {2024},
	pages = {10711--10732},
}

@misc{stoehr_localizing_2024,
	title = {Localizing {Paragraph} {Memorization} in {Language} {Models}},
	doi = {10.48550/arXiv.2403.19851},
	publisher = {arXiv},
	author = {Stoehr, Niklas and Gordon, Mitchell and Zhang, Chiyuan and Lewis, Owen},
	month = mar,
	year = {2024},
	note = {arXiv:2403.19851},
}

@misc{gao_pile_2020,
	title = {The {Pile}: {An} {800GB} {Dataset} of {Diverse} {Text} for {Language} {Modeling}},
	shorttitle = {The {Pile}},
	doi = {10.48550/arXiv.2101.00027},
	publisher = {arXiv},
	author = {Gao, Leo and Biderman, Stella and Black, Sid and Golding, Laurence and Hoppe, Travis and Foster, Charles and Phang, Jason and He, Horace and Thite, Anish and Nabeshima, Noa and Presser, Shawn and Leahy, Connor},
	month = dec,
	year = {2020},
	note = {arXiv:2101.00027},
}

@misc{springenberg_striving_2015,
	title = {Striving for {Simplicity}: {The} {All} {Convolutional} {Net}},
	shorttitle = {Striving for {Simplicity}},
	doi = {10.48550/arXiv.1412.6806},
	publisher = {arXiv},
	author = {Springenberg, Jost Tobias and Dosovitskiy, Alexey and Brox, Thomas and Riedmiller, Martin},
	month = apr,
	year = {2015},
	note = {arXiv:1412.6806},
}

@inproceedings{selvaraju_grad-cam_2017,
	title = {Grad-{CAM}: {Visual} {Explanations} {From} {Deep} {Networks} via {Gradient}-{Based} {Localization}},
	booktitle = {{IEEE} {ICCV}},
	author = {Selvaraju, Ramprasaath R. and Cogswell, Michael and Das, Abhishek and Vedantam, Ramakrishna and Parikh, Devi and Batra, Dhruv},
	month = oct,
	year = {2017},
}

@inproceedings{prashanth_recite_2024,
	title = {Recite, {Reconstruct}, {Recollect}: {Memorization} in {LMs} as a {Multifaceted} {Phenomenon}},
	shorttitle = {Recite, {Reconstruct}, {Recollect}},
	booktitle = {{ICLR}},
	author = {Prashanth, USVSN Sai and Deng, Alvin and O'Brien, Kyle and V, Jyothir S. and Khan, Mohammad Aflah and Borkar, Jaydeep and Choquette-Choo, Christopher A. and Fuehne, Jacob Ray and Biderman, Stella and Ke, Tracy and Lee, Katherine and Saphra, Naomi},
	month = oct,
	year = {2024},
}

@misc{zhang_extending_2025,
	title = {Extending {Memorization} {Dynamics} in {Pythia} {Models} from {Instance}-{Level} {Insights}},
	doi = {10.48550/arXiv.2506.12321},
	publisher = {arXiv},
	author = {Zhang, Jie and Zhao, Qinghua and Li, Lei and Lin, Chi-ho},
	month = jun,
	year = {2025},
	note = {arXiv:2506.12321},
}

@inproceedings{meng_locating_2022,
	title = {Locating and {Editing} {Factual} {Associations} in {GPT}},
	booktitle = {{NeurIPS}},
	author = {Meng, Kevin and Bau, David and Andonian, Alex and Belinkov, Yonatan},
	year = {2022},
}

@inproceedings{chen_multi-perspective_2024,
	title = {A {Multi}-{Perspective} {Analysis} of {Memorization} in {Large} {Language} {Models}},
	doi = {10.18653/v1/2024.emnlp-main.627},
	booktitle = {{EMNLP}},
	author = {Chen, Bowen and Han, Namgi and Miyao, Yusuke},
	year = {2024},
	pages = {11190--11209},
}

@inproceedings{menta_analyzing_2025,
	title = {Analyzing {Memorization} in {Large} {Language} {Models} through the {Lens} of {Model} {Attribution}},
	doi = {10.18653/v1/2025.naacl-long.535},
	booktitle = {{NAACL}:{HLT}},
	author = {Menta, Tarun Ram and Agrawal, Susmit and Agarwal, Chirag},
	year = {2025},
	pages = {10661--10689},
}

@inproceedings{dentan_predicting_2025,
	title = {Predicting {Memorization} {Within} {Large} {Language} {Models} {Fine}-{Tuned} for {Classification}},
	doi = {10.3233/FAIA251323},
	booktitle = {{ECAI}},
	author = {Dentan, Jérémie and Buscaldi, Davide and Shabou, Aymen and Vanier, Sonia},
	month = oct,
	year = {2025},
}
